\RequirePackage{rotating}
\documentclass[acmsmall,screen]{acmart}
\usepackage{multirow}
\usepackage{gensymb}
\usepackage{makecell}
\usepackage{graphicx}
\usepackage{subfigure}
\usepackage{caption}

\usepackage{amsmath, amsfonts, amssymb}
\usepackage{booktabs}
\usepackage{hyperref}
\usepackage{bm}
\usepackage{bbding}
\usepackage[doipre={doi:~}]{uri}
\AtBeginDocument{%
  }

\setcopyright{acmcopyright}
\copyrightyear{2023}
\acmYear{2023}
\acmDOI{XXXXXXX.XXXXXXX}

\acmJournal{JACM}
\acmVolume{37}
\acmNumber{4}
\acmArticle{111}
\acmMonth{7}

\begin{document}

\title{Spatial-Temporal Data Mining for Ocean Science: Data, Methodologies and Opportunities}

\author{Hanchen Yang}
\email{neoyang@tongji.edu.cn}
\orcid{0000-0002-9011-0355}
\affiliation{%
  \institution{Tongji University, The Hong Kong Polytechnic University}
  \streetaddress{4800 Caoan Rd}
  \city{Jiading Qu}
  \state{Shanghai Shi}
  \country{China}
  \postcode{200092}
}

\author{Wengen Li}
\authornote{Corresponding Author}
\email{lwengen@tongji.edu.cn}
\orcid{0000-0002-8768-6740}
\affiliation{%
  \institution{Tongji University}
  \streetaddress{4800 Caoan Rd}
  \city{Jiading Qu}
  \state{Shanghai Shi}
  \country{China}
  \postcode{200092}
}

\author{Shuyu Wang}
\email{tj_sywang@tongji.edu.cn}
\orcid{0000-0003-4731-6443}
\affiliation{%
  \institution{Tongji University }
  \streetaddress{4800 Caoan Rd}
  \city{Jiading Qu}
  \state{Shanghai Shi}
  \country{China}
  \postcode{200092}
}

\author{Hui Li}
\email{2230760@tongji.edu.cn}
\orcid{0009-0005-0763-4166}
\affiliation{%
 \institution{Tongji University }
 \streetaddress{4800 Caoan Rd}
 \city{Jiading Qu}
 \state{Shanghai Shi}
 \country{China}
 \postcode{200092}
}

\author{Jihong Guan}
\email{jhguan@tongji.edu.cn}
\orcid{0000-0003-2313-7635}
\affiliation{%
  \institution{Tongji University }
  \streetaddress{4800 Caoan Rd}
  \city{Jiading Qu}
  \state{Shanghai Shi}
  \country{China}
  \postcode{200092}
}

\author{Shuigeng Zhou}
\email{sgzhou@fudan.edu.cn}
\orcid{0000-0002-1949-2768}
\affiliation{%
  \institution{Fudan University}
  \streetaddress{4800 Caoan Rd}
  \city{YangPu Qu}
  \state{Shanghai Shi}
  \country{China}
  \postcode{200092}
}

\author{Jiannong Cao}
\email{jiannong.cao@polyu.edu.hk}
\orcid{0000-0002-2725-2529}
\affiliation{%
  \institution{The Hong Kong Polytechnic University}
  \streetaddress{Hung Hom}
  \city{Kowloon}
  \state{Hong Kong}
  \country{China}
}

\renewcommand{\shortauthors}{Hanchen Yang, Wengen Li and Jihong Guan et al.}

\begin{abstract}
With the rapid amassing of \textit{spatial-temporal}~(ST) ocean data, many \textit{spatial-temporal data mining} (STDM) studies have been conducted to address various oceanic issues, including climate forecasting and disaster warning. Compared with typical ST data (e.g., traffic data), ST ocean data is more complicated but with unique characteristics, e.g., diverse regionality and high sparsity. These characteristics make it difficult to design and train STDM models on ST ocean data. To the best of our knowledge, a comprehensive survey of existing studies remains missing in the literature, which hinders not only computer scientists from identifying the research issues in ocean data mining but also ocean scientists to apply advanced STDM techniques. In this paper, we provide a comprehensive survey of existing STDM studies for ocean science. Concretely, we first review the widely-used ST ocean datasets and highlight their unique characteristics. Then, typical ST ocean data quality enhancement techniques are explored. Next, we classify existing STDM studies in ocean science into four types of tasks, i.e., prediction, event detection, pattern mining, and anomaly detection, and elaborate on the techniques for these tasks. Finally, promising research opportunities are discussed. This survey can help scientists from both computer science and ocean science better understand the fundamental concepts, key techniques, and open challenges of STDM for ocean science.

\end{abstract}




\maketitle

\section{Introduction}

\begin{figure}[]
\begin{minipage}[t]{0.55\textwidth}
\centering
\includegraphics[width=\textwidth]{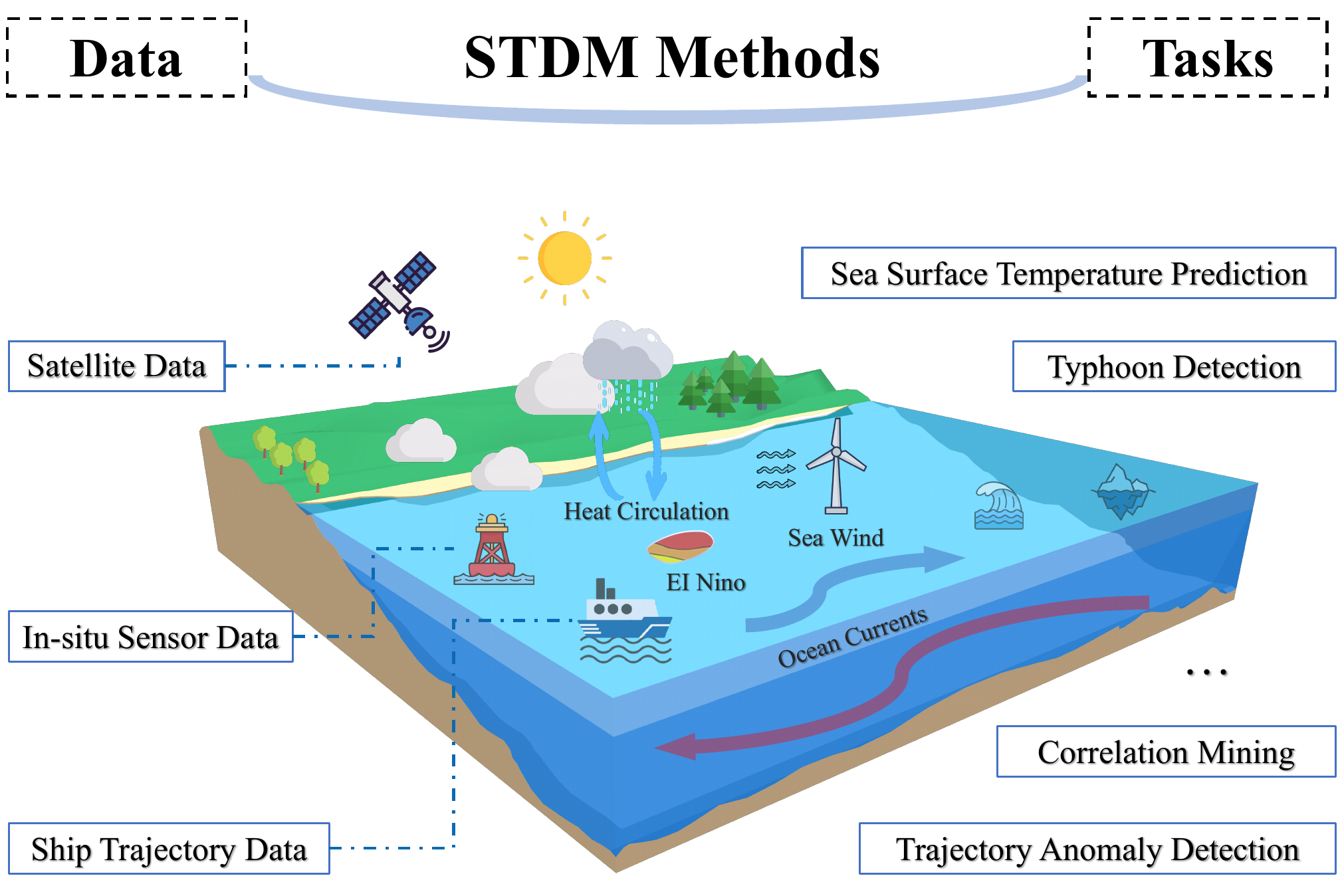}
\caption{The illustration of STDM for ocean science, which utilizes various STDM methods to discover unknown but potentially useful patterns from different ocean data sources for a number of data mining tasks.}
  \label{ocena}
\end{minipage}
\space{}
\begin{minipage}[t]{0.43\textwidth}
\centering
\includegraphics[width=\textwidth]{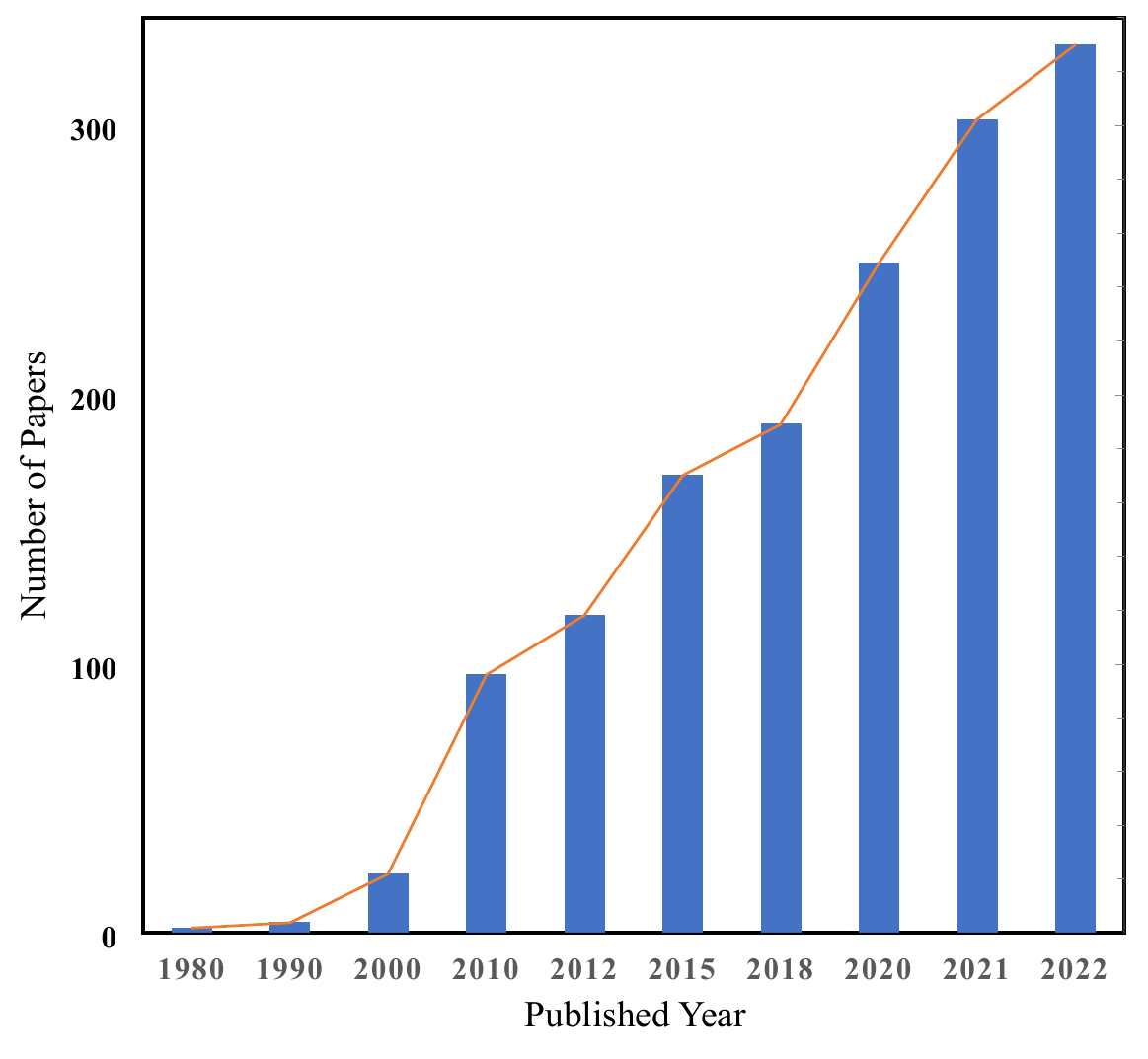}
\caption{1980-2022 publications of STDM studies for ocean science (Results from Web of Science with query keywords: [Spatial temporal \& ocean] or [Spatio-temporal \& ocean]).}
    \label{paper}
\end{minipage}
\end{figure}

Ocean, covering more than two-thirds of the earth, plays an important role in various applications (e.g., climate forecasting and ocean transportation~\cite{DBLP:journals/tai/LouLG23}), and is critical to human survival and sustainable development. 
Many real-world events and phenomena (e.g., El Ni$\mathrm{\tilde{n}}$o, typhoons and ocean currents) occurring in ocean generally not only are associated with spatial locations but also change over time.
Thus, the collected ocean data for monitoring these events and phenomena are typically spatial-temporal (ST) data in nature and require specific spatial-temporal data mining (STDM) techniques to conduct data analysis.
As shown in Fig.~\ref{ocena}, STDM for ocean science aims to uncover the ST patterns and correlations from various ST ocean data, helps us better understand the ocean system, and provides valuable support for various real-world applications. 
For example, accurate prediction of sea surface temperature~(SST) is of great significance in weather forecasting, El Ni$\mathrm{\tilde{n}}$o event detection, and disaster warning, and could also benefit aquaculture and agriculture~\cite{SVR}.

The past decades have witnessed ever-increasing ST ocean data and lots of research on STDM for ocean science, which lays the solid foundation for addressing various oceanic issues in the data-driven paradigm.
With the rapid development of data sensing technologies, numerous ocean observation sensors of different types have been deployed on various platforms including space remote satellites, automatic buoys, and vessels all over the world, producing large amounts of ST data of ocean. 
To be more specific, an estimate puts the size of ST ocean data in 2030 at 350 Petabytes~(1 PB = 1,000 TB), and this number is predicted to grow exponentially~\cite{overpeck2011climate}. 
With such huge amounts of ST ocean data, an increasing number of studies on STDM for ocean science have been conducted.
According to incomplete statistics of the related publications from 1980 to 2022 on Web of Science, as shown in Fig.~\ref{paper}, 
ocean STDM receives increasing attention over time and the increasing trend in the last three years is especially obvious. 
This trend indicates that STDM for ocean science has become a hot research topic in recent years. 

Though STDM for ocean science is a vibrant interdisciplinary field, there are still some obstacles that hinder its development and applications. 

First, there is a lack of review on available ST ocean datasets and their unique data characteristics.
In the past decades, ST ocean datasets collected from different sources have been released by various producers, e.g., National Oceanic and Atmospheric Administration~(NOAA), National Aeronautics and Space Administration~(NASA), and European Centre for Medium-Range Weather Forecasts~(ECMWF). However, these datasets have diverse data collection protocols, sensors, and technologies, leading to diversity in data resolution, coverage, and quality.
Currently, there is no systematic review of these data to facilitate researchers to choose appropriate datasets for studying different problems and tasks of ocean science.
Moreover, ST ocean data has some unique characteristics, e.g., diverse regionality, high sparsity, inherent uncertainty, and deep ST dependencies, which are also not well investigated.
For example, due to the large-scale coverage of the ocean on Earth, there exists diverse regionality in different areas. 
The collected ST ocean data in the Arctic Ocean may have opposite seasonality patterns from the data near the equator, which is difficult to model in a unified system.
According to \cite{lian2023gcnet},  the extreme weather in remote regions~(e.g., the Antarctica Ocean) makes it difficult to deploy the in-situ sensors to collect data, and the heavy cloud in some periods may cause more than 80\% of satellite data to be lost, which are two main reasons for the data sparsity problem.
The sensors' stability and transmission loss may both cause uncertainty in ocean data.
In addition, the complex and unknown ocean circulation as well as teleconnection between different regions in the ocean~\cite{li2021tropical} may bring deep and hidden ST dependencies in the data.
Ignoring these unique data characteristics could lead to low accuracy and bad interpretability of the STDM methods for ocean science.
Therefore, from the data perspective, it is necessary to analyze the widely-used ST ocean datasets and identify their unique data characteristics.


\begin{table*}
\centering
\caption{A summary of the related surveys on STDM.}
\label{tab-survey}
\renewcommand{\arraystretch}{1.3}
\resizebox{\linewidth}{!}{
\begin{tabular}{ccccccccc}
\hline
\multirow{4}{*}{Year} & \multirow{4}{*}{Ref.} & \multirow{4}{*}{Contribution}                                                                               & \multicolumn{6}{c}{Scope}                                                                                                                                       \\ \cline{4-9} 
                      &                       &                                                                                                             & \multirow{3}{*}{\makecell[c]{Data}} & \multirow{3}{*}{\makecell[c]{Data \\ Quality}} & \multicolumn{4}{c}{Tasks \& Methodologies}                                          \\ \cline{6-9} 
                      &                       &                                                                                                             &                                  &                                  & \makecell[c]{ST\\ Prediction}           & \multicolumn{1}{c}{\makecell[c]{Event\\ Detection} } & \multicolumn{1}{c}{\makecell[c]{Pattern\\ Mining} } & \makecell[c]{Anomaly\\ Detection}    \\ \hline
2014                  &   \makecell[c]{\cite{faghmousSpatiotemporal2014}}                      & \makecell[c]{Overview the STDM methods and   \\
give some case studies of climate data.} &                                  &                                  &          \color{blue}\checkmark            &          \color{blue}\checkmark            &           \color{blue}\checkmark           &                \\ \hline
2015                  &   \makecell[c]{\cite{shekharSpatiotemporal2015}}                      & \makecell[c]{Provide a complexity \\ comparison of typical STDM methods.  } &                                  &                                  &                   &          \color{blue}\checkmark            &           \color{blue}\checkmark           &                \\ \hline
2019                  &   \makecell[c]{\cite{wangDeep2022}}                      & \makecell[c]{Summarize the latest deep learning methods in STDM. } &                                  &                                  &      \color{blue}\checkmark            &                      &           \color{blue}\checkmark           &    \color{blue}\checkmark            \\ \hline
2020                  &   \makecell[c]{\cite{kim2020survey}}                      & \makecell[c]{Overview the  GAN-based models for ST data completion.  } &                                    &      \color{blue}\checkmark                             &                 &                      &                     &          \\ \hline
2021                  &   \makecell[c]{\cite{haghbinApplications2021}}                      & \makecell[c]{Overview the ST methods  for SST prediction.} &      \color{blue}\checkmark                              &                                  &          \color{blue}\checkmark            &                      &                    &                \\ \hline

2022                  &   \makecell[c]{\cite{wuMultisource2022}}                      & \makecell[c]{Summarize the current research methodologies \\ and the challenges of STDM in ocean domain.  } &  \color{blue}\checkmark                                    &                                  &      \color{blue}\checkmark            &          \color{blue}\checkmark            &                     &    \color{blue}\checkmark            \\ \hline
2022                  &   \makecell[c]{\cite{sharmaSpatiotemporal2022}}                      & \makecell[c]{Overview of the latest STDM methods for mutilple fieids.} &                                  &                                  &          \color{blue}\checkmark            &          \color{blue}\checkmark            &           \color{blue}\checkmark           &                \\ \hline
2022                  &   \makecell[c]{\cite{hamdi2022spatiotemporal}}                      & \makecell[c]{Summarize some challenging issues and \\ open problems of multiple STDM  directions.} &                                  &                                  &          \color{blue}\checkmark            &                     &           \color{blue}\checkmark           &   \color{blue}\checkmark              \\ \hline

\multicolumn{2}{c}{ \textbf{Ours} }                                       & \textbf{\makecell[c]{A survey of STDM methods for ocean science:\\ multi-source datasets with unique characteristics,\\ data quality enhancement methods, advanced\\ models in various tasks, and future directions.}} & \color{blue}\checkmark                                    &  \color{blue}\checkmark                                  &      \color{blue}\checkmark            &          \color{blue}\checkmark            &  \color{blue}\checkmark                    &   \color{blue}\checkmark        \\ \hline
\end{tabular}}
\end{table*}

Second, although there are already some surveys on STDM in the literature, we do not see any systematic and comprehensive surveys on STDM techniques for ocean science.
Table~\ref{tab-survey} summarizes and compares the related surveys of STDM in terms of contribution and scope. To provide a clear comparison, we put different oceanic STDM tasks into four types, i.e., ST prediction, event detection, pattern mining, and anomaly detection, which will be discussed in detail in Section~\ref{sec:tasks-methodologies}.
Concretely, Faghous et al.~\cite{faghmousSpatiotemporal2014} reviewed the STDM methods and conducted several case studies in the climate domain.
Ki et al.~\cite{shekharSpatiotemporal2015} provided an introduction to typical STDM methods from the perspective of computational complexity. 
Wang et al.~\cite{wangDeep2022} surveyed recent studies on deep learning techniques for STDM.
Kim et al.~\cite{kim2020survey} summarized the data completion methods for improving the quality of ST data for downstream applications.
Haghbin et al.~\cite{haghbinApplications2021} reviewed the datasets and different types of methods for sea surface temperature (SST) prediction. 
Wu et al.~\cite{wuMultisource2022} classified the STDM methods for ocean into statistic methods and machine learning methods, and gave case studies on some STDM tasks in ocean.
Sharma et al.~\cite{sharmaSpatiotemporal2022} analyzed the typical problems of STDM, with emphasis on the tasks of prediction and clustering for ST data.
Handi et al.~\cite{hamdi2022spatiotemporal} summarized some challenging issues in STDM, e.g., dynamic ST dependencies and poor data quality.
Although these studies above give introduction to some STDM methods related to ocean, there does not exist a survey that covers all the key components of STDM for ocean science, including available ST datasets, data processing methods to enhance data quality, STDM methods for various tasks, and future directions.
\begin{figure}[]
\begin{minipage}[t]{\textwidth}
\centering
\includegraphics[width=12.5cm]{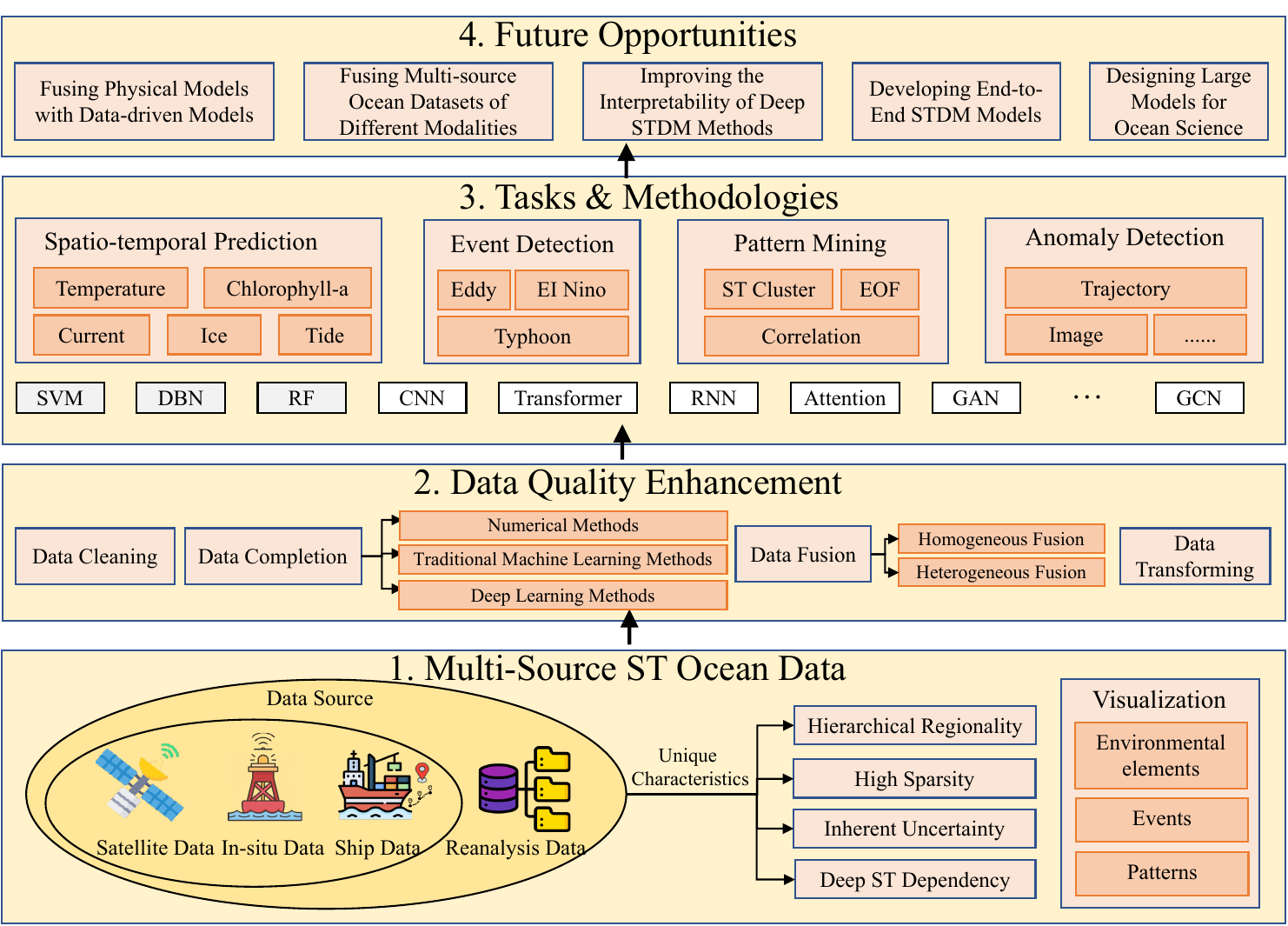}
\caption{The content and organization of this survey.}
  \label{frame}
\end{minipage}
\end{figure}

Third, STDM for ocean science is a typical multi-disciplinary research field, where rapidly evolving computer science meets conventional ocean science. The development of STDM for ocean science will not only help computer scientists identify new research issues,  but also assist the researchers of ocean science in applying advanced STDM methods to solving their problems, which could benefit various oceanic applications.
However, the existing and ever-arising challenges and opportunities of STDM for ocean science are not well studied.
For example, most existing methods only directly apply STDM methods to addressing ocean issues with large amounts of ST ocean data, ignoring the underlying physical laws of the ocean, which usually lack robustness and interpretability~\cite{haghbinApplications2021,wuMultisource2022}.
Thus, to further promote the development of STDM for ocean science, it is urgent to systemically and comprehensively summarize the related STDM studies for ocean science, and discuss their advantages and disadvantages to help researchers both in computer science and ocean science areas identify potential directions for further exploration.

Keep these in mind, in this paper, as shown in Fig.~\ref{frame}, we provide a comprehensive survey to systematically summarize the available ST ocean datasets, data quality enhancement methods, and STDM techniques for solving various ocean issues, and highlight the arising challenges. In summary, 
the contributions of this survey are fourfold as follows:

\begin{itemize}
\setlength{\itemsep}{0pt}
\setlength{\parsep}{0pt}
\setlength{\parskip}{0pt}
\item \textbf{Summarizing the wide-used and accessible ocean datasets.} We review different categories of ST ocean data, including satellite data, in-situ data, ship data, and reanalysis data, and present the popular datasets that are widely used from multiple dimensions (e.g., spatial resolution, temporal coverage, and related studies). In addition, we identify the unique data characteristics of ST ocean data and discuss typical data visualization methods.
\item \textbf{Reviewing the quality enhancement methods for ST ocean data.} We analyze the fundamental but necessarily important data quality enhancement methods, including data cleaning, data completion, data fusion, and data transforming for ST ocean data, and provide the corresponding representative models.
\item \textbf{Classifying the advanced STDM methods of typical ocean tasks.} 
This survey provides a comprehensive overview of the recent advances in STDM techniques for ocean science, and subsumes the main tasks into four types, i.e., ST prediction, event detection, pattern mining, and anomaly detection. For each type of these tasks, we provide a detailed review of the corresponding STDM techniques.
\item \textbf{Pinpointing the existing challenges and future directions.} We also identify the open problems that have not been well solved and provide some promising research directions in the future, which could help researchers in both computer science and ocean science identify new research topics and promote the development of smart ocean.
\end{itemize}

The rest of this survey is organized as follows:  
Section~\ref{sec:data} presents different categories of ST ocean datasets, identifies their unique characteristics, and provides typical ST ocean data visualization methods. 
Section~\ref{sec:quality} summarizes the techniques for improving the quality of ST ocean data, including data cleaning,  data completion, data fusion, and data transforming.
Section~\ref{sec:tasks-methodologies} classifies typical STDM tasks for ocean science into four types and presents a gallery of the technologies for each task. 
Section~\ref{sec:challenges} discusses the existing challenges and suggests future potential directions. We finally conclude this survey in Section 6.

\section{Data Sources, Unique Characteristics and Visualization}\label{sec:data}
In this section, as shown in Table~\ref{tab:comparison}, we first summarize four categories of ST ocean data, i.e., satellite data, in-situ data, ship data, and reanalysis data, according to the data sources, and  present the representative datasets with their basic information, including temporal periods, spatial resolution, spatial coverage, temporal resolution, and accessible storage. After that, we identify four unique data characteristics of ST ocean data, enabling researchers to understand the key difficulties in applying STDM techniques to solving ocean issues. Finally, the widely-used visualization methods for ST ocean data are discussed.

\subsection{Multi-source ST Ocean Data}
Typically, existing ST ocean data covers various ocean factors such as temperature, chlorophyll-a, ocean current, sea ice, and tide.
The majority of these data are collected by various measurement devices~(e.g., satellite sensors, in-situ sensors, and ship sensors) or from the outputs of reanalysis systems~(e.g., ECMWF reanalysis system and NOAA reanalysis system).
According to the data sources, ST ocean data can be divided into four categories, including satellite data, in-situ sensor data, ship data, and reanalysis data. 
Table~\ref{tab:comparison} briefly summarizes the advantages and disadvantages of the four categories of ST ocean data.
The satellite data and reanalysis data usually have wide spatial coverage and a long-term observation period but have low timeliness to obtain the original data.
The ship data and in-situ sensor data have better timeliness but lower data quality due to the data sparsity issue. In this section, we will introduce the typical and commonly used ST ocean datasets of the four data categories and summarize their features in Table~\ref{tab:performance-comparison}.

\begin{table}[]
 \caption{Comparison of different categories of ST ocean data.}
 \label{tab:comparison}
   \resizebox{0.9\textwidth}{!}{
 \centering
\begin{tabular}{lcccccc}
\hline
\textbf{Category} & \textbf{\makecell[c]{Spatial \\ coverage}} &  \textbf{\makecell[c]{Temporal \\ coverage}} & \textbf{\makecell[c]{Spatial \\ resolution}} &  \textbf{\makecell[c]{Temporal \\ resolution}} & \textbf{Quality} &  \textbf{Timeliness}\\ \hline
Satellite data & global     & long   & low    & days  & medium   & low         \\
In-situ data  & local    & short    & high  & hours  & medium   & high           \\
Ship data &  local  &  short  & high    & minutes & low   & medium    \\
Reanalysis data  & global      & long   & low    & days & high  & low          \\ \hline
	\end{tabular}}
\end{table}




\begin{sidewaystable}
  \caption{Public datasets and their features.}
 \centering
  \renewcommand{\arraystretch}{2.4}
  \resizebox{\textwidth}{!}{
  \label{tab:performance-comparison}
\begin{tabular}{c|cccccccc}
\hline
\textbf{Category}                   & \textbf{Name}        & \textbf{Period}           & {\textbf{ \makecell[c]{Spatial \\ Resolution}}} & {\textbf{Coverage}} & {\textbf{  \makecell[c]{Temporal \\ Resolution}}} & {\textbf{Citation}} & {\textbf{Type}}              & \textbf{Source} \\ \hline
\multirow{9}{*}{\textbf{Satellite Data}}  & \multirow{3}{*}{MODIS}       & 2000 to present           & 0.041x0.041                                      & Global                                 & 8 days                                            &       \makecell[c]{\cite{koner2019daytime,huang2021nonstationary,hussein2021spatiotemporal,doi:10.1080/20964471.2020.1776435}     }                            & \multirow{3}{*}{\makecell[c]{Sea surface temperature, ocean color, \\ sea surface salinity} }                      &  \multirow{3}{*}{\makecell[c]{\url{https://modis.gsfc.nasa.gov/}}}            \\
                                          &                              & 2002 to present           & 1 km x 1 km                                      & Global                                 & daily                                             & \makecell[c]{\cite{vazquez2019using}}                                         &                        &             \\
                                          &                              & 2002 to present           & 0.083\degree x 0.083\degree                                    & Global                                 & monthly                                           &  \makecell[c]{\cite{kilpatrick2015decade,kilpatrick2019alternating}      }                                    &                          &             \\
                                           & AVHRR                  & 1979 to present           & 1.1 km x 1.1 km                                  & Global                                 & daily                                            & \makecell[c]{ \cite{koner2019daytime,heCloudfree2003,hosodaGlobal2016} }                                      & Sea surface temperature, ocean color                                     & \makecell[c]{ \url{https://www.eumetsat.int/avhrr}}      \\                                       
                                          & Sentinel-3                   & 2016 to present           & 1.2 km x 1.2 km                                  & Global                                 & 5 days                                            & \makecell[c]{ \cite{DBLP:journals/lgrs/HuberDS23,DBLP:journals/aeog/BinhHTDT22,DBLP:journals/lgrs/Fernandez-Beltran22,DBLP:journals/remotesensing/LianXPLZYL22,DBLP:journals/tgrs/BoyCBT22,DBLP:journals/tgrs/XuL22,DBLP:journals/tgrs/LiuHLHIKYSHZW22} }                                      & Sea surface temperature, ocean color                                     & \makecell[c]{ \url{https://sentinels.copernicus.eu/web/sentinel/}}      \\
                                          & GOCI                         & 2010 to 2021              & 0.5 km x 0.5 km                                  & Korean sea                             & houtly                                            &  \makecell[c]{\cite{DBLP:journals/tgrs/YeomJHLLP22,DBLP:journals/remotesensing/ParkP21,DBLP:journals/remotesensing/SunSQWMH19,DBLP:journals/tgrs/LiuEWLLZ17}  }                                    & Sea surface chlorophy-ll, ocean color                       &  \makecell[c]{\url{https://oceancolor.gsfc.nasa.gov/data/goci/}}            \\
                                          & CZCS                         & 1978-1986                 & 0.825 km x 0.825 km                              & Global                                 & 8 days                                            &   \makecell[c]{\cite{DBLP:conf/icpr/ZhangHG96,DBLP:journals/tgrs/MukaiSMT92,DBLP:journals/lgrs/RaoARR05,DBLP:journals/tgrs/MelinCVH22} }                                    & Sea surface chlorophy-ll                        & \makecell[c]{\url{https://oceancolor.gsfc.nasa.gov/data/CZCS/}}             \\ 
                                          & OCM-2                        & 2009 to present           & 1-4 km x 1-4 km                                  & Global                                 & 2 days                                            & \makecell[c]{\cite{DBLP:journals/lgrs/RaoARR05}}                                       & Ocean color                                     & \makecell[c]{\url{https://ioccg.org/sensor/ocm-2/}}            \\
                                          & SeaWIFS                      & 1997-2010                 & 1-4 km x 1-4 km                                  & Global                                 & daily                                             & \makecell[c]{\cite{DBLP:journals/tgrs/MelinCVH22,DBLP:journals/staeors/WangGL19,DBLP:journals/lgrs/LiXLOZ15,DBLP:journals/lgrs/ChenG13}     }                                  & Ocean color                                     & \makecell[c]{ \url{https://oceancolor.gsfc.nasa.gov/SeaWiFS/}}            \\ \hline
\multirow{3}{*}{\textbf{In-situ data}}         & Argo                         &  1996-present                         &   Trajectories of  about 14060 floats                                               & Global                                    & 1-10 days                                                  &     \makecell[c]{ \cite{DBLP:journals/tgrs/BanksGSS12,DBLP:journals/lgrs/BhaskarJ15,DBLP:journals/remotesensing/ChenGGXLGG22,DBLP:journals/gandc/XueZXS22,DBLP:journals/aeog/XiaoTLCYXLH22}}                               &                 sea surface temperature, sea surface salinity                                   & \makecell[c]{\url{https://argo.ucsd.edu/}}    \\              
                                          & GO-BGC                       &   2021-present                        &  Trajectories  of about 500 floats                                              &   Global                                     & 10 days                                                    &     \makecell[c]{ \cite{DBLP:journals/remotesensing/XingBZC20,DBLP:journals/remotesensing/ShuXXQMBC22,DBLP:conf/chinacom/CaiLZ19,DBLP:journals/tgrs/HuCBHLP23}}                             &     Sea $O_{2}$, sea Ph                                               &  \makecell[c]{\url{https://www.go-bgc.org/}}            \\
                                          & SOCCOM                       &  2004-present                         &       Trajectories of about 200 floats                                          &       Antarctic Ocean                                 &        10 days                                           &     \makecell[c]{\cite{https://doi.org/10.1029/2022JC018859,AcousticFloatTrackingwiththeKalmanSmoother}   }                                &  Ocean carbon                                                 &   \makecell[c]{\url{https://soccom.princeton.edu/} }     \\ \hline
\multirow{2}{*}{\textbf{Ships Data}}      & AIS          &  2016 to 2018 &   Trajectories of about 70,000 vessels                                                 &  Global                                     & 30 seconds - 1 day                                        &   \makecell[c]{\cite{DBLP:journals/mis/HuangZHWR21,DBLP:conf/igarss/ElvidgeGHZ22,DBLP:journals/eswa/Galotto-TebarPC22,DBLP:journals/geoinformatica/BrandoliRSABPRR22} }                                    & Trajectory anomalies, ship tracking                            & \makecell[c]{\url{https://www.vmsdata.com/}}            \\
                                          & VMS       & April, 2020               &  Trajectories of 750,000 vessels                                                &  Global                                      & 30 seconds - 1 day                                                    & \makecell[c]{ \cite{DBLP:conf/bigdataconf/ShahirTGCZW19,DBLP:journals/cgf/Storm-FurruB20,DBLP:journals/remotesensing/GuanZZLMLBC21}}                                      & Trajectory anomalies                           &  \makecell[c]{\url{https://marinecadastre.gov/}}
                                                    \\ \hline
\multirow{8}{*}{\textbf{Reanalysis Data}} & OISST                        & 1979-present              & 0.25\degree x0.25\degree                                        & Global                                 & daily                                             &  \makecell[c]{\cite{DBLP:journals/remotesensing/ZhuLWL22,MGCN,DBLP:journals/lgrs/XieOZJSX22,DBLP:journals/lgrs/JahanbakhtXA22}  }                                   & Sea surface temperature                         &\makecell[c]{\url{https://www.ncei.noaa.gov}}
            \\
                                          & ERA-5                        & 1959-present                 & 4\degree x4\degree                                              & Global                                 & 12-hour                                           &  \makecell[c]{\cite{DBLP:journals/tgrs/ZhangLHL22,  DBLP:journals/lgrs/McNichollLCD22, DBLP:journals/remotesensing/LiuZZ23, DBLP:journals/remotesensing/KuchinskaiaBPNDKSPBBSKZ23} }                                    & Sea surface temperature                         & \makecell[c]{\url{https://www.ecmwf.int}}          \\
                                          & CMEMS Level 3 SLA            & 2004  to present          & 0.125\degree x0.125\degree                                      & Global                                 & daily                                             & \makecell[c]{\cite{DBLP:conf/igarss/AoufHTC18, DBLP:journals/sensors/GbagirC20}}                                       & Sea level anomalies                             & \makecell[c]{\url{https://marine.copernicus.eu/}}          \\
                                          & CMEMS                        & 1993-2020                 & 0.25\degree x0.25\degree                                        & Global                                 & daily                                             & \makecell[c]{\cite{DBLP:conf/igarss/AoufHTC18}}                                      & Sea surface height anomaly                 & \makecell[c]{\url{https://marine.copernicus.eu/}}         \\
                                          & HadCRUT4                     & 1961-1990                 & 5\degree x5\degree                                              & Global                                 & monthly                                           & \makecell[c]{\cite{cowtan2014coverage,morice2012quantifying,schurer2018interpretations} }                                      & Air/Marine temperature anomalies                & \makecell[c]{\url{https://www.metoffice.gov.uk/}}  \\
                                          & COBE SST & 1891 to present           & 1\degree x1\degree                                              & Global                                 & monthly                                           &  \makecell[c]{\cite{chen2020long,hausfather2017assessing}}                                      & Sea surface temperature                         & \makecell[c]{\url{https://psl.noaa.gov/data/}}          \\
                                          & COBE-SST 2 and Sea Ice       & 1850  to 2019             & 1\degree x1\degree                                              & Global                                 & monthly                                           &  \makecell[c]{\cite{tang2022analysis}, \cite{chen2020long}}                                      & Sea surface temperature, sea ice concentration & \makecell[c]{\url{https://psl.noaa.gov/data/}} \\
                                          & CMAP Precipitation           & 1979 the present.         & 2.5\degree x2.5\degree                                          & Global                                 & monthly                                           & \makecell[c]{\cite{li2015decadal,dong2023effect,kim2019intercomparison}}                                     & Pentad global gridded precipitation means.      & \makecell[c]{\url{https://psl.noaa.gov/data/}}   \\
                                          & WOD           & 1772 to 2017.         & 1\degree  x1\degree                                          & Global                                 & daily                                           & \makecell[c]{\cite{boyer2013world}}                                     &  Sea temperature, salinity, oxygen     & \makecell[c]{\url{https://www.ncei.noaa.gov/products/world-ocean-database}}  
                                        \\ \hline
\end{tabular}}
\end{sidewaystable}

\begin{figure*}[]
\centering
\subfigure[The collection of original satellite data.]{
\includegraphics[width=0.46\textwidth]{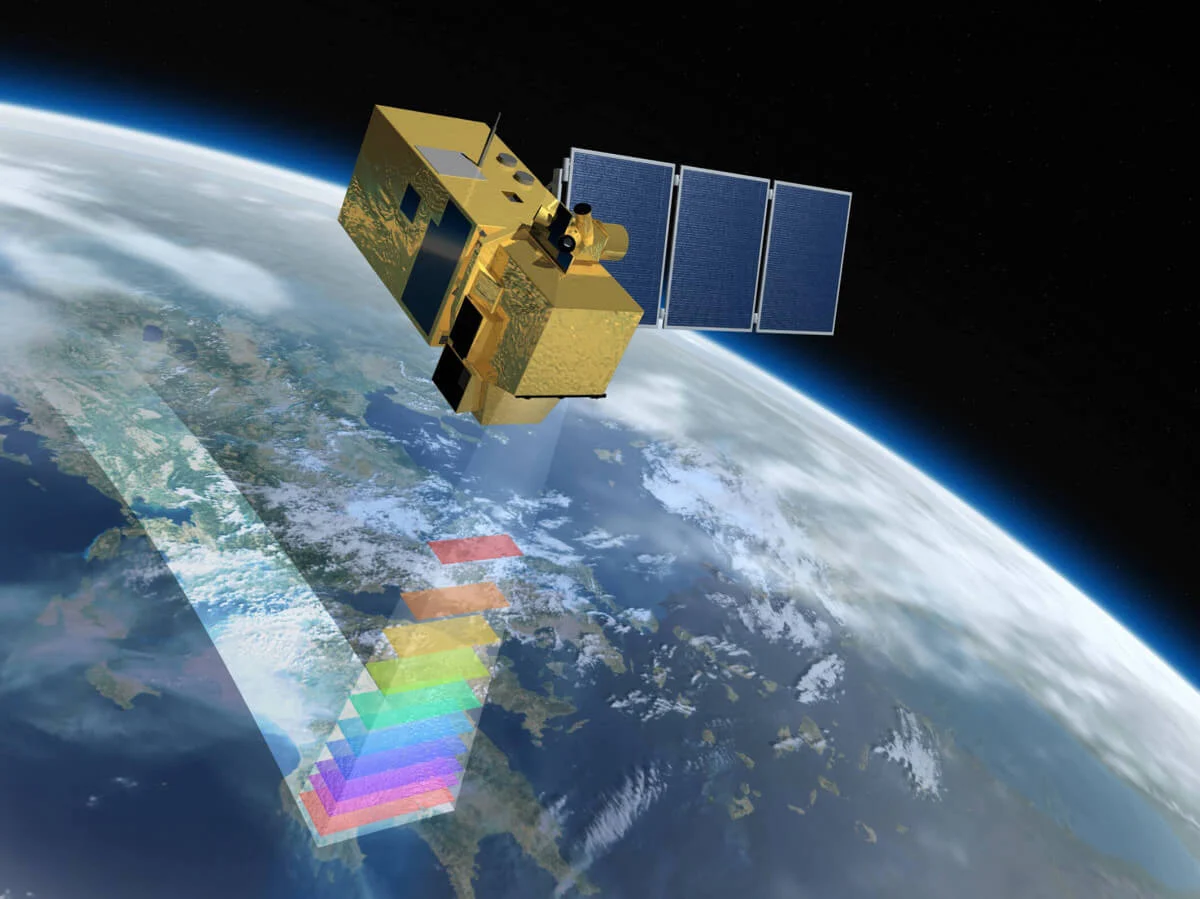}
\label{satellite1_a}
}
\subfigure[The inversion process of satellite data.]{
\includegraphics[width=0.5\textwidth]{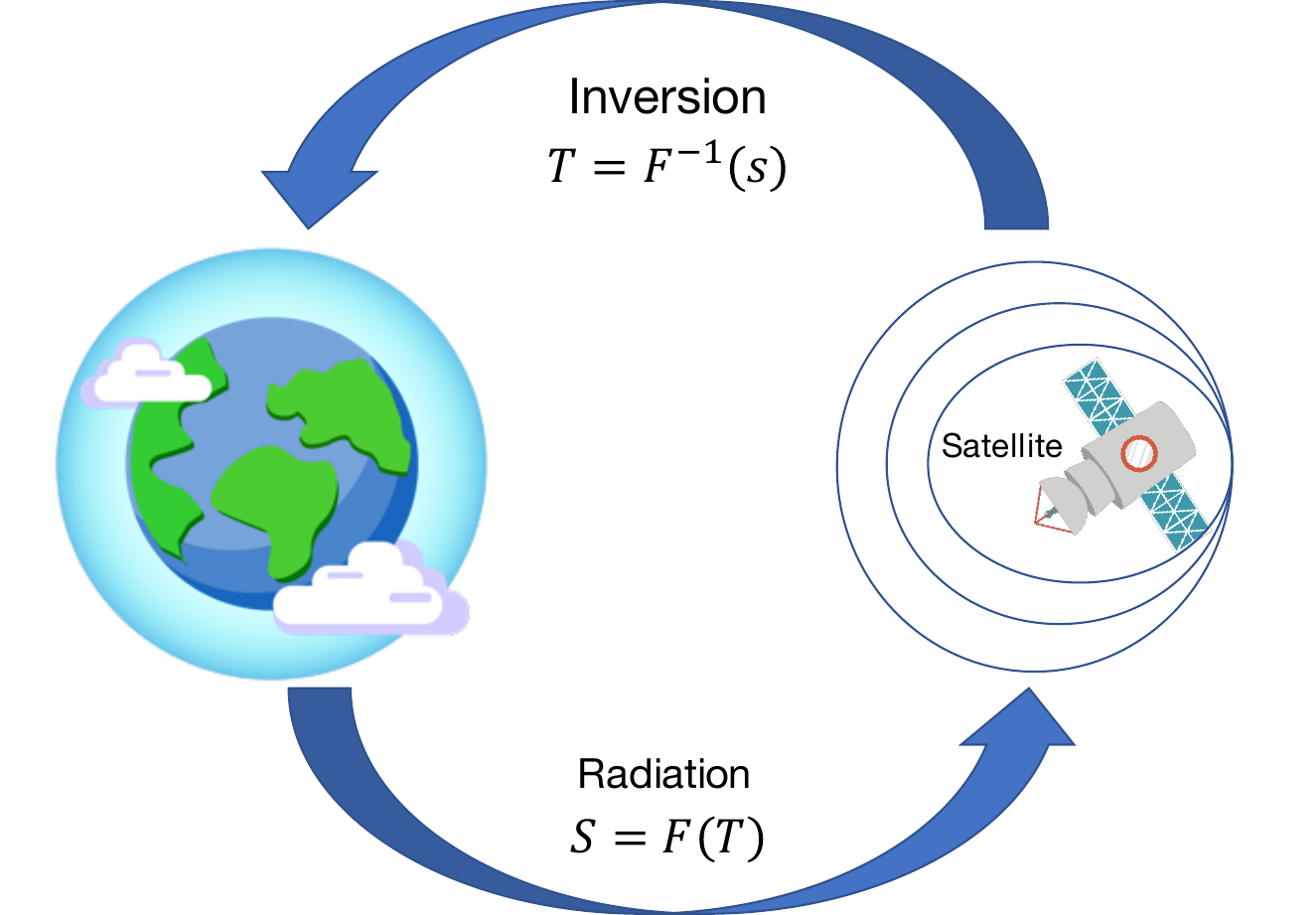}
\label{satellite1_b}
}
\caption{Satellite data collection and inversion.  (a) Satellite data collection with different bands~\cite{Pic1Label}; (b) Data inversion to process the original satellite data.}
\label{satellite1}
\end{figure*}

\subsubsection{Satellite Data}
Satellite data, also named remote sensing data, became available in the late 1960s and are a source of high-quality data for monitoring the ocean~\cite{liu2006using}.
The view from space allows ocean factors ~(e.g., SST, and sea surface chlorophyll-a) to be measured globally, especially in remote regions where is not convenient to deploy sensors.
Additionally, satellites can ensure large spatial coverage and frequent observation, even in extreme weather conditions. 
During the past decade, much progress has been made in ocean monitoring using satellites, and the quality of satellite instruments and the accuracy of inversion algorithms have improved considerably, making satellite observations be widely used in various ocean tasks, e.g., sea ice prediction~\cite{zakhvatkina2019satellite} and typhoon detection~\cite{ruttgers2018typhoon}.




As shown in Fig.~\ref{satellite1_a}, satellite data collection is to detect the thermal energy emitted by the earth's surface using spaceborne microwave or infrared radiometers~\cite{jimenez2003generalized}. 
The surface radiation is modified in its passage through the atmosphere to the radiometer by a bunch of processes, e.g., atmospheric absorption, emission, and scattering.
Thus, with appropriate data inversion, we can obtain valid estimates of various ocean factors. 
The general satellite data inversion process is shown in Fig.~\ref{satellite1_b}, where $S$ denotes the radiation signal emitted by the ocean surface, and $T$ is the target data obtained by the data inversion function $F$. The data inversion function $F$ is generally non-linear, its inverse function is $F^{-1}$.
Through the data inversion algorithms~\cite{liu2006using,zhang2009study, SHI201816,li2011regional}, the radiation signal can be transferred to ST time series data of various ocean factors, e.g., SST and sea surface chlorophyll-a. 

Many countries have launched satellites for various ocean applications and we introduce the wide-used satellite data as below.


\begin{itemize}
\setlength{\itemsep}{0pt}
\setlength{\parsep}{0pt}
\setlength{\parskip}{0pt}
\item \textbf{MODIS:} The Moderate Resolution Imaging Spectroradiometer~(MODIS) is a key instrument onboard the Earth Observing System~(EOS) Terra and Aqua platforms, and designed to monitor the atmosphere, ocean, and land surface. MODIS has a viewing swath width of 2,330 km and views the entire surface of the Earth every two days. It measures 36 spectral bands between 0.405 and 14.385 $\upmu$m, and acquires data at three spatial resolutions, i.e., 0.25km, 0.5km, and 1km.
\item \textbf{AVHRR:} The Advanced Very High-Resolution Radiometer~(AVHRR) multi-purpose imaging instrument is used for monitoring global cloud cover, sea surface temperature, ice, snow, and vegetation cover characteristics. AVHRR provides four-to-six-band multispectral data from the NOAA polar-orbiting satellite series. AVHRR provides continuous global coverage since June 1979 and its spatial resolution is 1.1 km.
\item \textbf{Sentinel-3:} Sentinel-3 is a dedicated Copernicus satellite mission delivering a variety of high-quality ocean and atmosphere measurements.
The main objective of the mission is to collect parameters such as sea surface topography, sea surface temperature, and ocean surface color. It provides two-day global coverage optical data with two satellites and altimetry measurements for sea and land applications with real-time data products delivered in less than three hours. The spatial resolution of Sentinel-3 is 1.2 km.
\item \textbf{GOCI:} Geostationary Ocean Color Imager (GOCI) is the first ocean color sensor launched to monitor ocean color around the Korean Peninsula. 
GOCI has a large temporal coverage and it has a revisit time of around 1 hour.
The spatial resolution of GOCI is about 0.5 km.
\item \textbf{CZCS:} The Coastal Zone Color Scanner (CZCS) is a multi-channel scanning radiometer aboard the ocean remote sensing satellite Nimbus 7. CZCS could map chlorophyll concentration, sediment distribution, salinity, ocean currents, and the temperature of coastal waters. CZCS measures reflect solar energy in six channels at a spatial resolution of 0.8 km.
\item \textbf{SeaWIFS:} Sea-viewing Wide Field-of-view Sensor~(SeaWiFS) is loaded on the OrbView-2 satellite launched by NASA in 1997. SeaWIFS aims to obtain accurate ocean color data for the global ocean and makes this data readily available to researchers. SeaWiFS has 6 watercolor bands with a bandwidth of 20nm, i.e., 412nm, 443nm, 490nm, 510nm, 555nm, and 670nm. Compared with CZCS, the performance indicators of this sensor have been greatly improved, such as more reasonable band settings, higher signal-to-noise ratio, and richer band information. SeaWiFS has two spatial resolutions, i.e., 1.1 km and 4.5 km, and its temporal resolution is daily or once every two days.
\end{itemize}

\subsubsection{In-situ Sensor Data} 
In-situ sensor data refers to the data collected where the instrument is physically located.
It is a traditional but important way to collect data for ocean factors and has a long history since the 1600s~\cite{faghmousSpatiotemporal2014}. 
In the past few decades, numerous in-situ sensors of different types, e.g., buoys sensors, underwater sensors, and station sensors, have been deployed all over the world. However, considering the wide spatial coverage of the ocean, in-situ sensor data are still sparse in space and time since they are only available when and where the sensors are physically located. 
In addition, due to the data privacy issue, there is only a small portion of in-situ sensor data available to the public.
We introduce the three available and widely used in-situ sensor datasets, i.e., Argo, Go-BGC, and SOCCOM, as below.

\begin{figure}[]
\begin{minipage}[t]{0.55\textwidth}
\centering
\includegraphics[width=\textwidth]{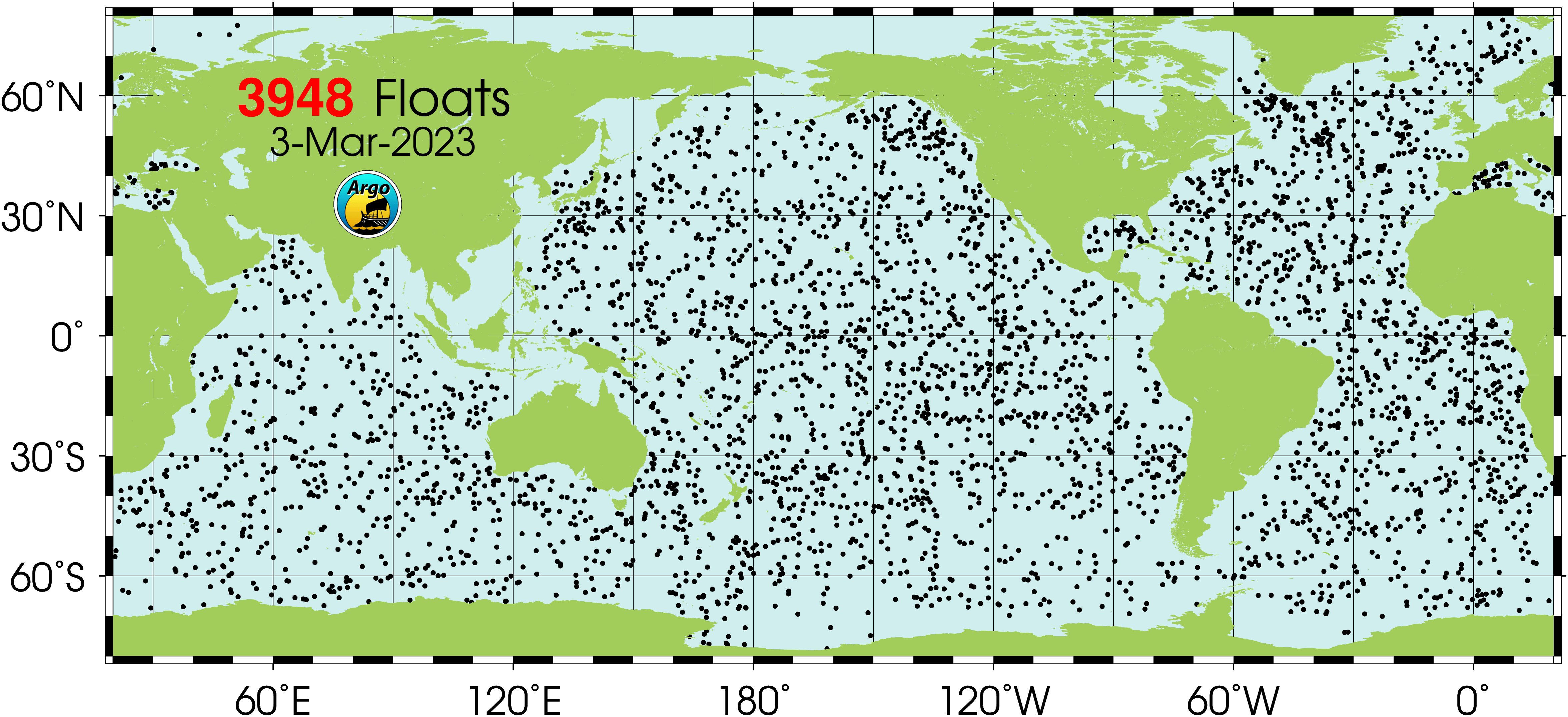}
\caption{The global distribution of Argo sensors in 2023~\cite{Pic3Label}.}
   \label{sensor}
\end{minipage}
\space{}
\begin{minipage}[t]{0.4\textwidth}
\centering
\includegraphics[width=\textwidth]{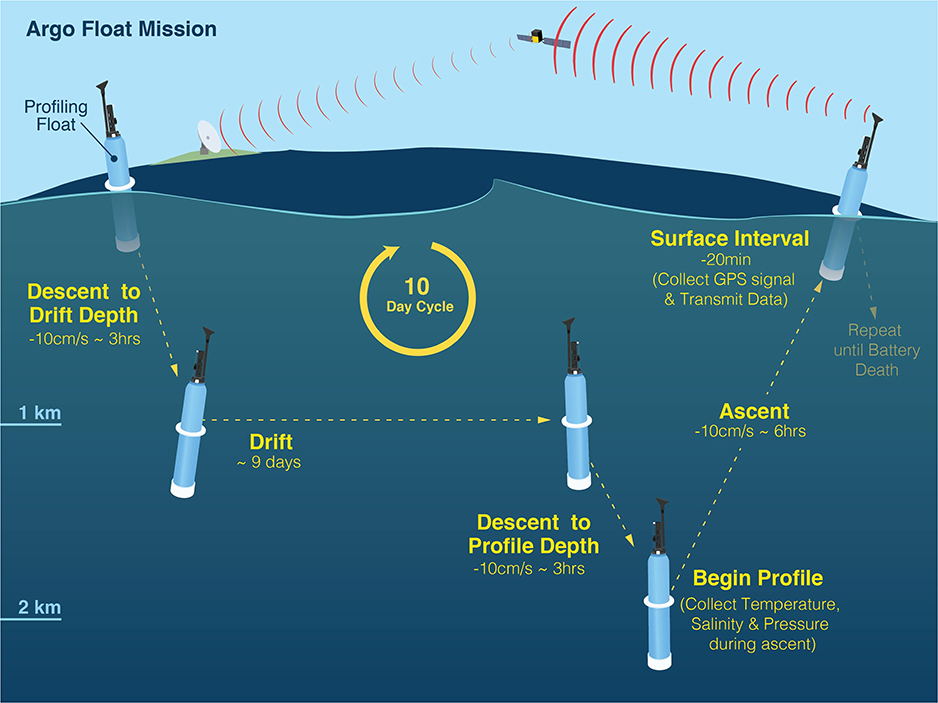}
\caption{The workflow of Argo sensors~\cite{Pic4Label}. }
  \label{sensorwork}
\end{minipage}
\end{figure}


\begin{itemize}
\setlength{\itemsep}{0pt}
\setlength{\parsep}{0pt}
\setlength{\parskip}{0pt}
\item \textbf{Argo:} Array for Real-time Geostrophic Oceanography~(Argo) is an international program that uses profiling floats to observe temperature, salinity, currents, and other factors in the ocean. It has been in operation since the early 2000s and provides real-time data for studying climate and oceanography. Fig.~\ref{sensor} shows the global distribution of 3948 sensors from the Argo program~\cite{Pic3Label}. As shown in Fig.~\ref{sensorwork}, the Argo float mission is a 10-day cycle, and Argo sensors drift at a depth of 1000 meters (the so-called parking depth) and measure the conductivity, the temperature, etc, as it moves back up to the ocean surface. Once the float is on the surface, it will communicate with a satellite and send its location and collected data to them, then continue its mission~(around 4-5 years).
\item \textbf{GO-BGC:} The Global Ocean Biogeochemistry (GO-BGC) Array is a project to build a global network of chemical and biological sensors on profiling floats. The network monitors biogeochemical cycles and ocean health, including the elemental cycles of carbon, oxygen, and nitrogen through all seasons of the year. 
\item \textbf{SOCCOM:} The Southern Ocean Carbon and Climate Observations and Modeling project (SOCCOM) is a multi-institutional program focusing on unlocking the mysteries of the Antica Ocean, exploring the influence of carbon, nutrients, oxygen, and other factors on global climate.
\end{itemize}


\begin{figure}[]
\space{}
\begin{minipage}[t]{0.9\textwidth}
\centering
\includegraphics[width=0.8\textwidth]{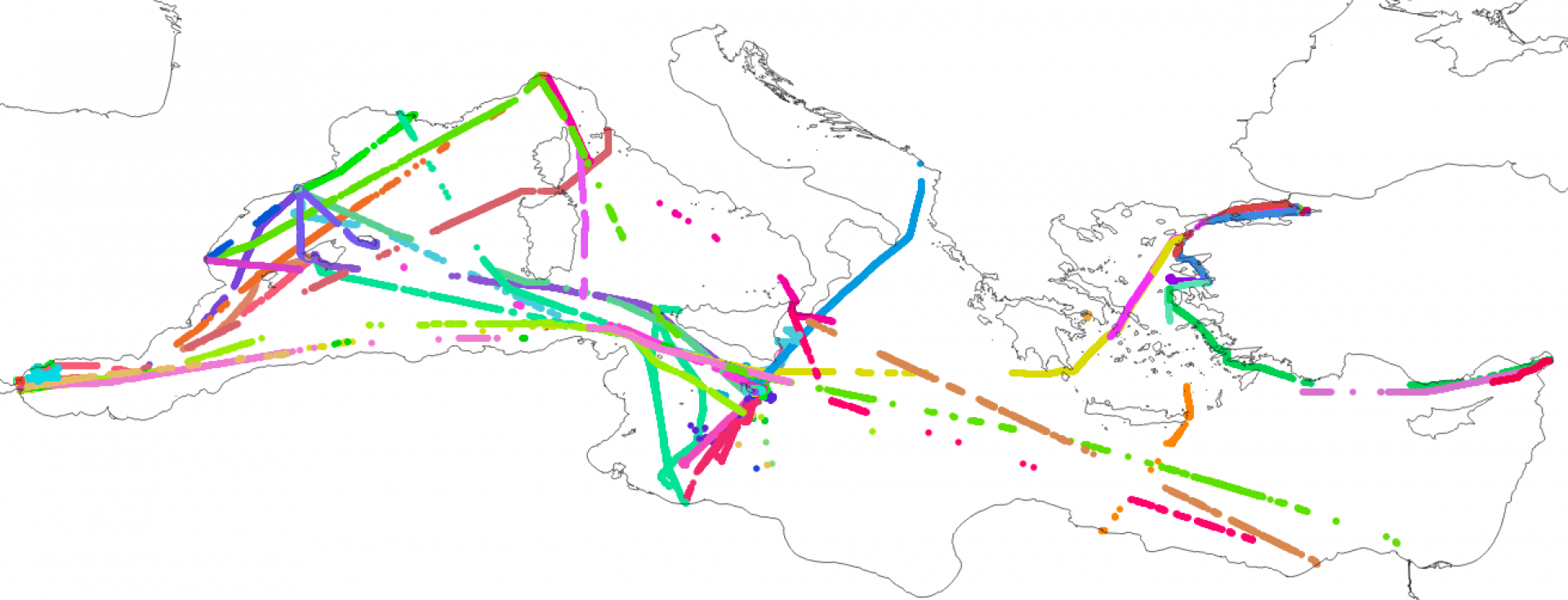}
\caption{The visualization of ship trajectories in the Automatic Identification System~(AIS)~\cite{Pic8Label}. }
  \label{AIS}
\end{minipage}
\end{figure}

\subsubsection{Ship Data}
Ship data is collected by ships and vessels for directly observing the ocean factors, collecting biology samples, and investigating the ocean~\cite{DBLP:journals/cgf/Storm-FurruB20}. Ships can collect marine data at different locations along their trajectories. Here, we introduce two common tracking systems that record the information of ships.
 
\begin{itemize}
\setlength{\itemsep}{0pt}
\setlength{\parsep}{0pt}
\setlength{\parskip}{0pt}
\item \textbf{AIS:} Automatic identification system~(AIS) is an operation tracking system that provides ships' information such as the type of ship, location, and speed. Fig.~\ref{AIS} visualizes several AIS trajectories of ships.
AIS was used by the U.S. Coast Guard in 2009 for the first time to monitor and transmit the locations of huge vessels. According to International Maritime Organization (IMO), the goals of developing AIS are to help recognize ships, assist in target tracking, enhance the safety of marine traffic, and make the flow of vessel traffic smooth. Since 2004, installing AIS transponders for vessels over 300 tons of weight is mandatory in the International Convention for Safety of Life at Sea (SOLAS) guideline.
\item \textbf{VMS:} Vessel Monitoring System~(VMS) is a form of satellite tracking using transmitters on board fishing vessels. It tracks vessels in a similar way to AIS but is restricted to government regulators or other fisheries authorities. All European Union, Faroese, and Norwegian vessels which exceed 12 meters overall length must be fitted with VMS units.
VMS collects the locations, times, and oceanic features, along the trajectories of vessels.
\end{itemize}

\subsubsection{Reanalysis Data} 
It is important to have complete observations of the ocean system in both spatial and temporal dimensions. 
However, the observations discussed above are usually unevenly distributed.
Even in satellite data, the observations cannot provide a complete recording of the state of the ocean system across the globe at a given point in time. 
Reanalysis data is the most completed data, which fills the gaps in the observational records through the simulation models or the physic models~\cite{DBLP:journals/remotesensing/ZhuLWL22}.
Reanalysis datasets are often globally complete and consistent in time, and referred to as `maps without gaps'. 
They are among the most-used datasets in ocean, especially in various data-driven ocean applications. 
We introduce the two widely-used reanalysis datasets, i.e., ERA5 and OISST, as follows:

\begin{itemize}
\setlength{\itemsep}{0pt}
\setlength{\parsep}{0pt}
\setlength{\parskip}{0pt}
\item \textbf{ERA5:} ECMWF Reanalysis Atmosphere V5~(ERA5) is published by ECMWF for monitoring the global climate, which provides hourly data with estimates of uncertainty for the parameters of atmospheric, land-surface, and ocean. ERA5 data are available in the Climate Data Store on regular latitude-longitude grids at 0.25\degree x0.25\degree \ resolution, providing atmospheric parameters on 37 pressure levels, sea surface temperature, sea wave height, etc.  
\item \textbf{OISST:} NOAA's Optimum Interpolation Sea Surface Temperature (OISST, also known as Reynolds' SST) is a series of global reanalysis products, including the weekly OISST data on 1\degree x1\degree \ grids and daily OISST data on 0.25\degree x0.25\degree \ grids. OISST generates smoothed and complete ocean observation data by conducting interpolating and extrapolating operations based on a combination of ocean temperature observations from satellites and in-situ platforms (e.g., ships and buoys).
The input data is first mapped to specific regular grids since they are usually irregularly distributed in space. Then, statistical methods, e.g., optimum interpolation (OI)~\cite{huangImprovements2021a}, are applied to fill in the missing values.
\end{itemize}

%

\subsection{Unique Data Characteristics}
Compared with typical ST data~(e.g., traffic flow data and epidemic outbreak data), ST ocean data not only contains ST dependencies between adjacent locations but also has a lot of complex and unique characteristics.
Here, we summarize four unique characteristics of ST ocean data as below.

\textbf{Diverse Regionality.} 
In general, the ocean can be divided into a lot of regions from different aspects.
For example, one general partition is based on the climate zones to divide the ocean into various regions, such as tropical zones and frigid zones. On the one hand, each region has diverse patterns from others. For example, the collected ocean data in the Arctic Ocean may have opposite patterns from the data near the equator~\cite{10.1145/3597937}. On the other hand, when learning the ST correlations in ocean, it is of high importance to capture the ST correlations between regions since the ocean is a unified system.
Moreover, due to the ocean current and atmospheric circulation, the correlations between regions are not static but dynamically changing over time.
This regionality brings difficulties for one model to learn the diverse features of all ocean regions, thus requiring adaptive learning for different regions.

\textbf{High Sparsity.} Sparsity is an important characteristic of ST ocean data.
For in-situ data and ship data, they are collected in a sparse and uneven distribution in both space and time, which leads to high sparsity in observation data for certain periods and some ocean areas. For satellite data, the collected data may be highly missed due to the cloud covering and the scanning cycle of satellites, and the sensors' stability and transmission loss.
For example, according to~\cite{lian2023gcnet}, the missing rate of MODIS satellite data could be higher than 80\% in some periods.
As a result, it is difficult to obtain continuous, accurate, and uniform ST ocean data in a wide range. In this case, STDM models have to learn knowledge and patterns from incomplete data, which makes the overfitting problem more serious. Thus, how to address the sparsity issue in ST ocean data is an important yet challenging task for developing STDM methods for ocean science.

\textbf{Inherent Uncertainty.} Another characteristic of ST ocean data is the inherent uncertainty due to the instability of the data collection process, which may cause the collected data to contain noise and deviate from real values.
Concretely, the uncertainty in ST ocean data stems from the fact that many ocean datasets have biases in sampling and measurement. 
For instance, uncertainty might occur because a particular thermometer is miscalibrated or poorly sited. 
However, data producers seldom provide uncertain information about data. 
For example, one may have access to three different SST datasets: one reanalysis dataset at 2.5\degree x 2.5\degree \ spatial resolution, another dataset at 0.75\degree x 0.75\degree \ spatial resolution, and a satellite dataset at 0.25\degree x 0.25\degree \ spatial resolution. 
Given that each dataset has its own biases, how to effectively fuse these datasets to obtain the correct information remains a challenging issue.

\textbf{Deep Spatial-temporal Dependency.} 
With the wide spatial coverage of the global ocean and long-term temporal record, ST ocean data has more complex and deep ST dependencies than typical ST data. 
Concretely, from the spatial view, the wide spatial coverage brings complicated ST dependencies between different ocean locations.
For example, the well-known El Ni$\mathrm{\tilde{n}}$o event, occurring in the equatorial Pacific Ocean, has been proven to cause intense storms and extreme temperatures in many regions far away from the Pacific Ocean via the tropical tropospheric and atmospheric bridge mechanism~\cite{EINO02}.
From the temporal view, ST ocean data may have long-term temporal dependencies and short-term dependencies, which requires STDM methods to have the capability to model such hierarchical temporal dependencies.
For example, SST shows a certain cyclical pattern in the year, and the SST in winter is usually differently distributed from that in summer.
In addition, the spatial dependency and temporal dependency in ST ocean data are often entangled, which makes learning the ST dependencies a non-trivial task.

\subsection{Data Visualization}
Visualization of ocean data is important for scientists and engineers to understand various ocean phenomena and discover their underlying complex and dynamic patterns. 
Since interactive visual analysis integrates the experts' knowledge into the design of the models and algorithms~\cite{liVROcean2011}, it allows us to interactively explore, analyze and assess data at different spatial-temporal scales to gain more insights into the data than traditional ways of data analysis. 
We introduce three typical types of visualization methods for ST ocean data as below.

\begin{itemize}
\setlength{\itemsep}{0pt}
\setlength{\parsep}{0pt}
\setlength{\parskip}{0pt}
\item \textbf{Ocean environmental elements visualization.}
Ocean environmental elements are a number of basic physical properties that affect the ocean ecosystem, including scalar data (e.g., temperature and salinity) and flow fields (e.g., trajectories and ocean currents).
2D plots (e.g., points, lines, surfaces, particles, and glyphs) and 3D volume rendering are widely-used methods for the visualization of these ocean environmental elements.
For example, Fig.~\ref{Visualization1} visualizes the global SST at different depths and the distribution of global surface ocean pH.
Currently, with the rapid development of hardware technology, the GPU-based hybrid graphics visualization approach is popular~\cite{liVROcean2011}. 
Su et al.~\cite{suMultidimensional2016} developed an ocean data visualization system that supports line contouring, volume rendering, and dynamic simulation of the sea current field. 
The system employs GPU-based rendering to visualize scalar or flow fields efficiently and is used for monitoring the changing processes of ocean environmental elements. 
Liu and Chen~\cite{liuFramework2017} also developed a framework for interactive visual analysis of heterogeneous marine data. 
Ocean environmental elements visualization could help us well understand the inherent operating processes of the ocean system.

\begin{figure*}[]
\centering
\subfigure[The visualization of global SST at different depths.]{
\includegraphics[width=0.5\textwidth, height=0.25\textwidth]{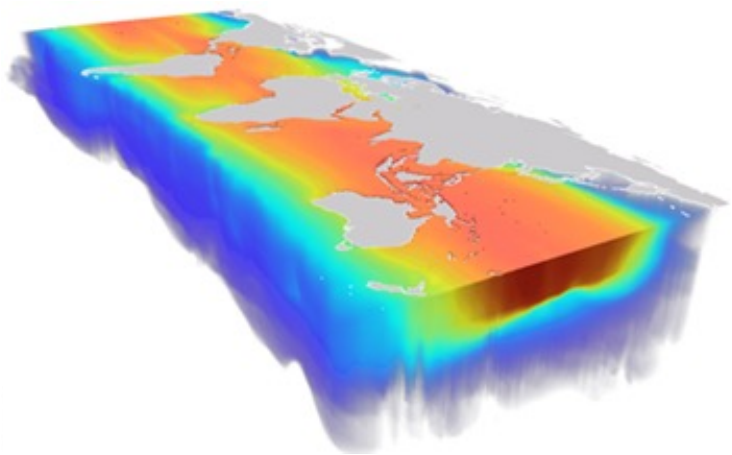}
}
\subfigure[The visualization of global surface ocean pH.]{
\includegraphics[width=0.46\textwidth, height=0.25\textwidth]{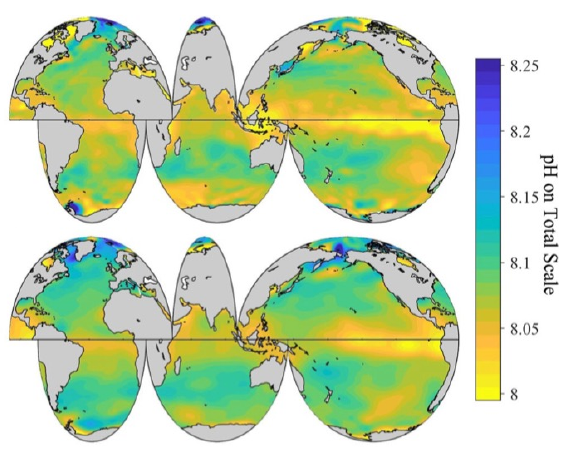}
}
\caption{The visualizations of ocean environmental elements: (a) The global SST at different depths~\cite {Pic5Label}, and  (b) the global surface ocean pH from two different analysis projects~\cite{jiang2019surface}.}
\label{Visualization1}
\end{figure*}

\begin{figure}[]
\begin{minipage}[t]{0.48\textwidth}
\centering
\includegraphics[width=0.9\textwidth, height=0.5\textwidth]{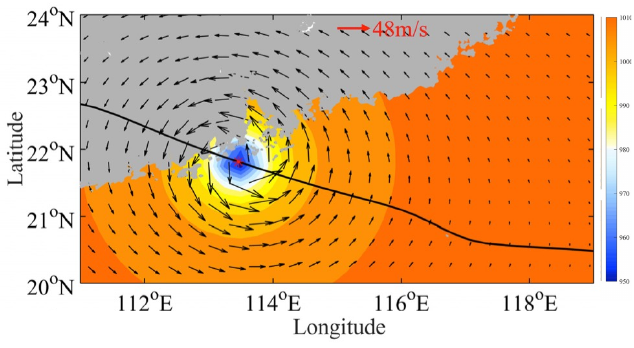}
\caption{The visualization of typhoon Hato at 04:00 UTC on 23 August 2017, where the black line shows the path of Hato~\cite{du2020effects}. }
   \label{typhoon}
\end{minipage}
\space{}
\begin{minipage}[t]{0.48\textwidth}
\centering
\includegraphics[width=0.9\textwidth, height=0.5\textwidth]{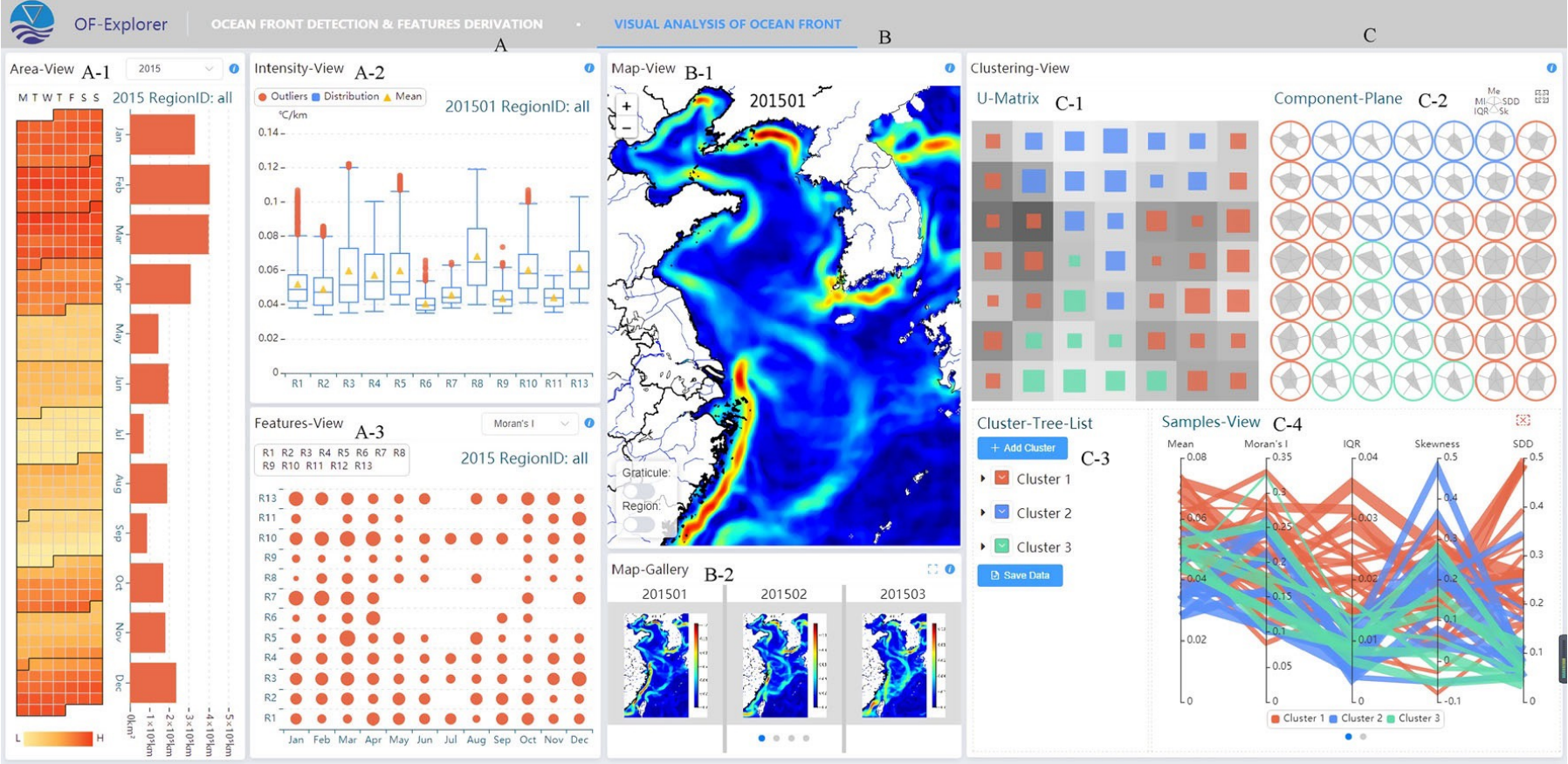}
\caption{The interface of  OFExplorer, a visualization tool for exploring and discovering SST characteristics and patterns~\cite{xie2022ofexplorer}.}
  \label{sd}
\end{minipage}
\end{figure}

\item \textbf{Ocean event visualization.} 
The interaction between various ocean environmental elements constitutes a variety of ocean events.
For instance, ocean eddies, ocean fronts, typhoons, and EI Ni$\mathrm{\tilde{n}}$o are common events occurring in the ocean, which often last for several days to several months.
Each event has different characters and needs special algorithms for visualization.
Various methods have been proposed to visualize these events. For example. Fig.~\ref{typhoon} visualizes the trajectory of typhoon Hato in 2017. 
George et al.~\cite{georgeInteractive2014} proposed a framework for the visual analysis of the sea level rise event, and provided multiple visual analytic tools for interactive hydrodynamic flux calculation on spatial-temporal and multivariate data. 
Hollt et al.\cite{holltOvis2014} developed an interactive visualization system for sea surface height data to support interactive exploration and analysis of the spatial distribution and the temporal evolution of ocean eddies. 
In general, the spatial and temporal scales of ocean events are often different, which requires event visualization methods to be adaptable to changeable scales~\cite{doi:10.1177/1473871613481692}. 



\item \textbf{Ocean pattern visualization.} 
Ocean patterns are related to complex ocean processes, which are hidden in the spatial and temporal variability of ocean data.
ST pattern visualization is used for understanding the evolution characteristics of the ocean environmental elements (e.g., temperature and salinity) or phenomena (e.g., eddies and fronts) over time to detect trends, periodicity, and other important factors. 
For example, Xie et al.~\cite{xie2022ofexplorer} developed the visualization tool OFExplorer, as shown in Fig.~\ref{sd}, to discover the changing patterns of SST. 
Another ocean pattern visualization approach integrates the changes of ocean environmental elements in both time and space into a 3D volume rendering \cite{reschWebbased2014} by reducing the feature dimension through slicing, averaging, and aggregation operations, or using PCA-like methods~\cite{sachaVisual2017, sipsVisual2012}. 
\end{itemize}

\section{Data Quality Enhancement} \label{sec:quality}
High data quality is of great significance to ensure that the results of STDM are reliable, accurate, and complete~\cite{kim2020survey}. 
However, in practice, it is difficult to get ST ocean data with high quality for many reasons such as the limited spatial and temporal coverage, the failures of data collection devices, and the loss during data transmission and storage. 
Meanwhile, the diversity of the data sources and the big volume of ST ocean data also seriously affect the quality of ST ocean data.
Therefore, to address the data quality issue, a number of studies have been conducted on data quality enhancement, which can make ST ocean data more suitable for specific applications. 
In this section, we overview three foundational data processing operations, i.e., data cleaning, data completion, and data fusion, and the corresponding techniques to improve the quality of ST ocean data. 
In addition, we introduce the data transforming process, which aims to transform the data representation to meet the specific requirements of the STDM tasks.

\subsection{Data Cleaning}
Data cleaning aims to remove incorrectly formatted data which may result in inaccurate analysis and unreliable results. Data cleaning requires a good understanding of the real distributions and statistical implications of the original data. Existing methods, e.g., \textbf{statistical analysis methods}, \textbf{proximity measurement methods} and \textbf{density-based clustering methods}, for data cleaning usually use constraints, rules, and patterns to detect and remove the outliers in data~\cite{zhou2020knowledge}. 
Statistical analysis methods~\cite{dasu2012statistical} detect the outliers that do not follow the given data distributions or regression equations. Proximity measurement methods first define a proximity measure between data points and then identify those abnormal data points that are far away from the other data points.
Density-based clustering methods~\cite{duan2009cluster} detect the data outliers by comparing their local density with the neighboring data points and are suitable for non-uniformly distributed data.
For the ST data in ocean, we utilize the above methods to remove the incorrectly formatted data points and improve the data quality for downstream STDM tasks.

\begin{figure*}[]
\centering
\subfigure[The spatial distribution of SST records from AVHRR satellite data on 3 October 2021, where the white areas have no data, and the gray areas are lands~\cite{Pic6Label}.]{
\includegraphics[width=0.47\textwidth]{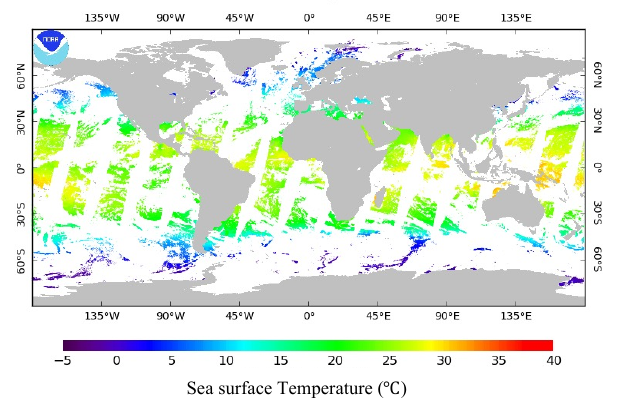}
\label{fig_sst_missing}
}
\space{}
\subfigure[The spatial distribution of Chl-a records from MODIS satellite data on 3 October 2021, where the black areas have no data~\cite{Pic7Label}.]{
\includegraphics[width=0.47\textwidth]{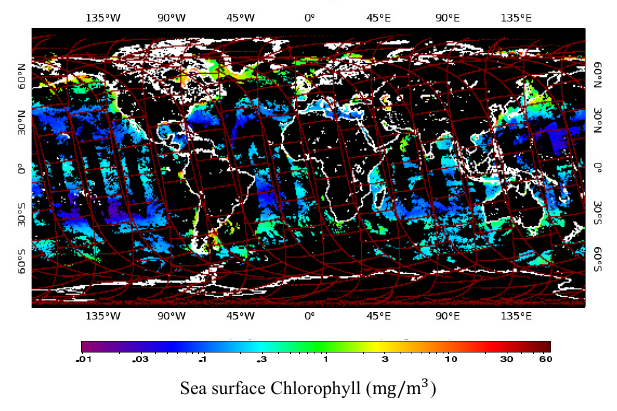}
\label{fig_cha_missing}
}
\subfigure[The monthly missing percentages of SST and Chl-a in the area of Northern South China Sea from 1997 to 2018~\cite{maTwoDecade2021}.]{
\includegraphics[width=0.98\textwidth,height=0.3\textwidth]{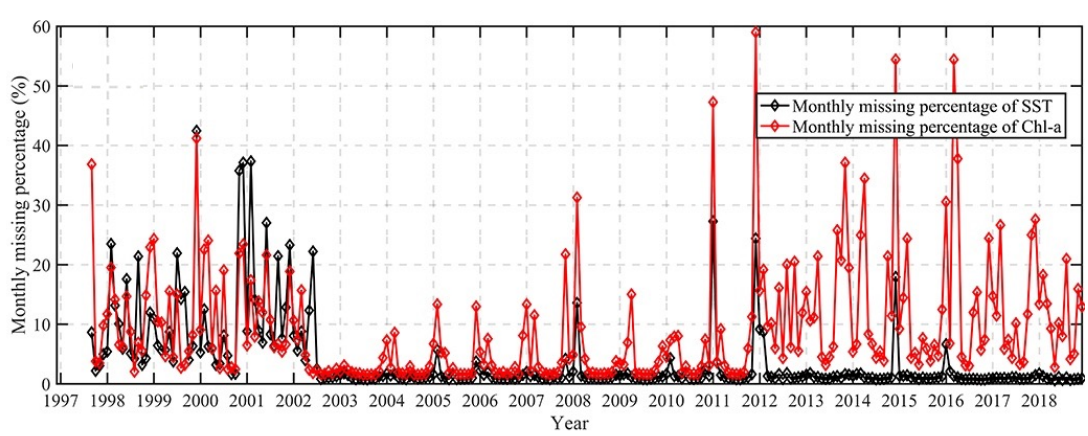}
\label{fig_missing_over_time}
}
\caption{Distributions and missing rates of global SST and Chl-a data.  (a) / (b) The spatial distribution of SST / Chl-a records from AVHRR / MODIS satellite data on 3 October 2021, and (c) the monthly average missing rates of SST and Chl-a in the Northern South China Sea from 1997 to 2018. 
}
\label{data2}
\end{figure*}


\subsection{Data Completion}
The issue of data missing or sparsity is very common in ST ocean data due to the influence of various inevitable factors such as cloud occlusion, bad weather, and sensor failure.  
If missing values are not well treated, the data analysis results may be unreliable and inaccurate, leading to bias in further applications. 
As shown in Fig.~\ref{fig_sst_missing} and Fig.~\ref{fig_cha_missing}, due to the cloud occlusion, both AVHRR data and MODIS data have many missing values. 
According to Fig.~\ref{fig_missing_over_time}, the monthly missing rate of MODIS data in the Northern South China Sea is often between 20\% and 50\% and the missing rate of SST even reaches 60\% in 2012, let alone the daily data missing rates.
Apparently, compared with typical ST data such as traffic data and crowd volume data, the missing rate in ocean data is extremely high, making it difficult to conduct data analysis and restricting the development of real-world applications~(e.g., weather forecasting and typhoon detection).
Therefore, data completion is regarded as a necessary step in ST ocean data quality enhancement and a variety of data completion methods have been proposed. 
In this section, we briefly introduce the problem definition of data completion and the related techniques. 

\begin{figure*}[]
  \centering
\includegraphics[width=\textwidth]{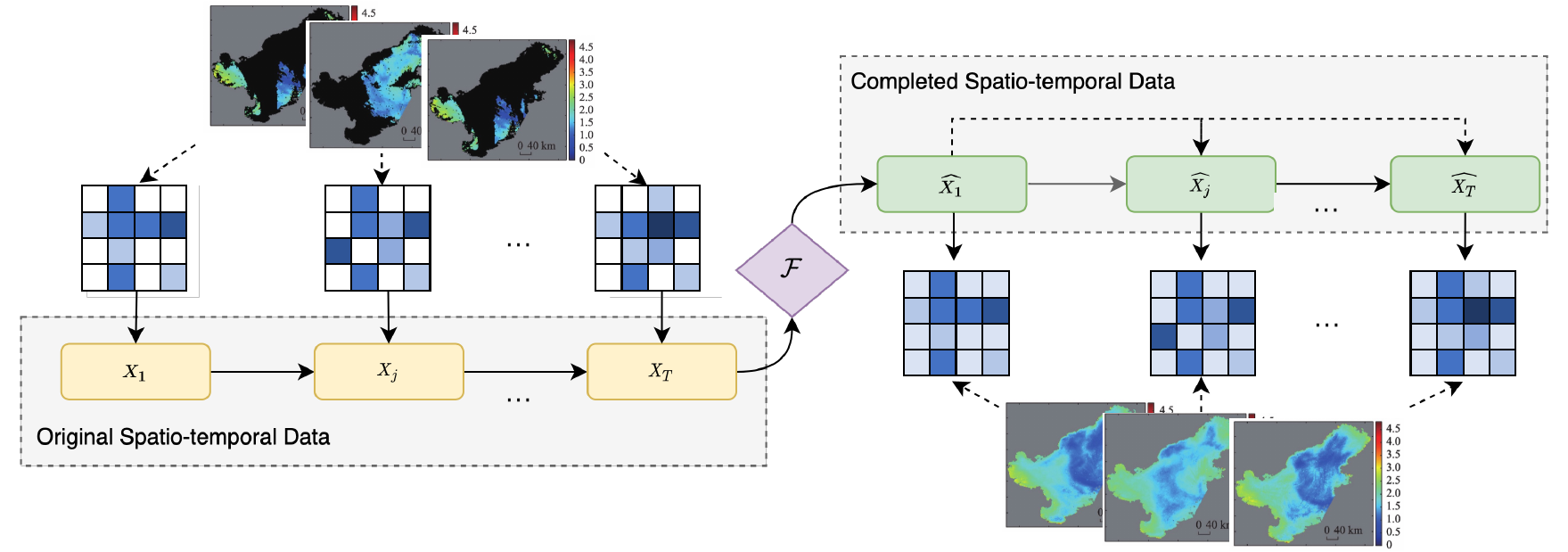}
  \caption{The general framework of ST data completion, where the input is incomplete ST data, the output is completed data, and $\mathcal{F}$ could be various data completion algorithms.}
    \label{framework of data completion}
\end{figure*}

As shown in Fig.~\ref{framework of data completion}, we use matrix $X=(X_1,..., X_j,..., X_T)\in {{\mathbb{R}}^{N\times T}}$ to represent the time-series of recorded ocean data for $N$ nodes (e.g., grid ocean regions and sensor locations) at $T$ timestamps, where $X_j$ represents the observation values for all nodes at timestamp $\tau_j$, and $x_{i,j}$ is the observation value of node $v_i$ at timestamp $\tau_j$. 
In addition, a masked matrix $M=(M_1,...,M_j,...,M_T)\textbf{}\in {{\mathbb{R}}^{N\times T}}$ is introduced for $X$, where each entry $m_{i,j}=0$ if the entry $x_{i,j}\in X$ is missing, otherwise $m_{i,j}=1$.

Given the incomplete data matrix $X$ and its masked matrix $M$, the data completion problem aims to fill the missing values in $X$ and ensure that the filled values are close to the real values, i.e., 

\begin{equation}
\label{3}
{{\widehat{X}}} = \bm {\mathcal{F}_\theta}\{{{X}}, {M}\},
\end{equation}
\begin{equation}
\label{4}
{{\widetilde{X}}}={{X}}+{{\widehat{X}}}\odot \left( {\mathrm O}-M \right),
\end{equation}
where ${{\widehat{X}}}$ is the generated data matrix by data completion model function $\bm {\mathcal{F}}$, \bm{$\theta$} denotes all the learnable parameters in the model, ${\mathrm O}$ is a matrix whose entries are all ones, and $\odot $ denotes Hadamard product. Finally, we can get the completed data ${{\widetilde{X}}}$.

Existing ST data completion methods can be roughly divided into three categories, i.e., numerical methods, traditional machine learning-based methods, and deep learning methods. 
And the summarization and comparison of these methods are in Table~\ref{com}.

\begin{table*}
\centering
\caption{Major existing methods of ocean data completion.}
\label{com}
\resizebox{\linewidth}{!}{
\begin{tabular}{cllcll}
\toprule
\textbf{Category}   & \textbf{Strengths} & \textbf{Weaknesses} & \textbf{Model}    & \textbf{Year} & \textbf{Approach}                \\
\midrule
\multirow{4}{*}{Numerical methods} &   \multirow{4}{*}{\makecell[l]{Need a small amount of data, \\ simple, fast, flexible, and adaptable.}}  & \multirow{4}{*}{\makecell[l]{Cannot model non-linear \\ data, depend on a correct \\ domain knowledge. }}  & OI~\cite{huangImprovements2021a}            & 1987 & min-variance                                    \\

   &  & & Kriging ~\cite{oliver1990kriging}    & 1990 & linear combination of nearby data                  \\
    & &  & IDW~\cite{bartier1996multivariate} & 1996 &  averaging neighbors' data   \\
  &  & & DINEOF~\cite{alvera2005reconstruction}             & 2005 & Empirical Orthogonal Function (EOF)                    \\
\midrule
\multirow{7}{*}{\makecell[c]{Traditional machine learning \\ based methods }}  &  \multirow{7}{*}{\makecell[l]{Can model dynamic, non-linear, \\ and noisy data, \\ easy to set up, self-adaptable, \\ self-organizing, and error tolerant.}}  &   \multirow{7}{*}{\makecell[l]{Have a high computational \\ cost and tend to overfit \\ when applied  to \\ long-term data.}} & RF~\cite{parkData2020}     & 2006 &  a multitude of decision trees               \\
   & & & XGBoost~\cite{shenDMAD2018}     & 2014 & gradient boosting decision trees  \\
     & &  & ANN~\cite{pisoniArtificial2008}   & 2015 & artificial neural network                \\                
  & & & KNN~\cite{huMachine2021}     & 2013 & averaging k-nearest neighbors            \\
               &  &   & SOM~\cite{jouiniReconstruction2013}     & 2013 & self-organizing map \\ 
       &  & & SVR~\cite{mohebzadehMachine2021}     & 2005 & calculating the best model to fit points           \\
            &  &   & MF~\cite{liuMissing2022}     & 2006 & matrix decomposition           \\
\midrule
\multirow{6}{*}{Deep learning methods}  &  \multirow{6}{*}{\makecell[l]{Extract deep features from \\ the data, can work with noisy data, \\ captures temporal dependencies \\ over variable periods of time, \\ high accuracy. }}          &   \multirow{6}{*}{\makecell[l]{High complexity, \\ high computational cost, \\ need to be tuned carefully \\ lack of interpretability }}     & NN~\cite{jamet2012retrieval}            & 2012 & neural network                                 \\

   & & & CNN~\cite{vientDataDriven2021}       & 2020 &  convolutional neural network              \\
   & & & DINCAE~\cite{jouiniReconstruction2013}       & 2016 & convolutional auto-encoder         \\
   & & & GAN~\cite{pan2021tec}       & 2021 & generative adversarial network         \\
   &   &   & GCN~\cite{lian2023gcnet}       & 2022 & graph completion network         \\
      &   &   & STA-GAN~\cite{wang2022sta}       & 2023 & generative adversarial network         \\
\bottomrule
\end{tabular}}
\end{table*}

\subsubsection{Numerical methods}
Numerical methods learn the linear correlations from past observations and the domain knowledge ~(e.g., physical laws of the ocean) to achieve data completion. 
History average~(HA) only uses the average of past data to fill in the missing values, which is too simple to learn the non-linear relationships.
To simulate the true data distribution, the Optimal Interpolation~(OI) method \cite{huangImprovements2021a} uses the min-variance to obtain the ideal unbiased approximation of missing data and is widely used to build cloud-free SST products by fusing data from multiple platforms, such as satellite data, in-situ data, and ship data. For example, the OI method is employed in the NOAA OISST products~\cite{huangImprovements2021a} and the blended foundation SST products~\cite{hosodaGlobal2016}.
In addition, the OI method is popular for analyzing daily SST by fusing the AVHRR satellite data and the Tropical Rainfall Measuring Mission Microwave Imager products~\cite{heCloudfree2003}. 
Distance-based data completion methods fill the miss data with the average of the data within a certain physical distance and are also popular in the early ages. For example, Kriging~\cite{bhattacharjeeSpatial2014, luAdaptive2008} and IDW \cite{shenDMAD2018}) achieve good performance on SST data completion based on the spatial relevance of original data. 
Beckers proposed the DINEOF method~\cite{beckersEOF2003} to achieve missing value completion in oceanographic data based on the Empirical Orthogonal Function~(EOF). 
DINEOF is widely used for the completion of Chl-a data~\cite{liuGap2018, guoVariability2022}, SST data~\cite{liReconstruction2021, maTwoDecade2021}, ocean wind data~\cite{zhaoInterpretation2010} and multivariate ocean factors~\cite{alvera-azcarateMultivariate2007}. 


In practice, numerical methods are simple and flexible, and widely used for filling in missing values in various ocean data products.
However, they cannot well learn the complex nonlinear relationships in ocean data, and some methods are highly dependent on expert knowledge
to set the values of parameters, which is not sufficient to achieve accurate ST data completion.

\subsubsection{Traditional machine learning-based methods}
Traditional machine learning-based methods aim to utilize historical ocean data to train machine learning models to simulate data distributions for data completion.
Various machine learning methods, e.g., Random Forest (RF), eXtreme Gradient Boosting (XGBoost), and Matrix Factorization (MF),  have been applied for ST ocean data completion~\cite{huMachine2021, mohebzadehMachine2021, parkReconstruction2019}.
Shen et al.~\cite{shenDMAD2018} utilize the RF method to fill the missing values in satellite data by constructing multiple decision trees. 
Chen et al.~\cite{chenImproving2019} used RF to improve the coverage of Chl-a data with two external factors, i.e., the Ocean Color Index and the Rayleigh-corrected Reflectivity. 
Hu et al.~\cite{huMachine2021} proposed an RF method with XGBoost to complete the sea surface Chl-a data.
KNN and SVR have been also adopted to fill missing values of ocean data based on distance measurement~\cite{sunderMachine2020, mohebzadehMachine2021, huMachine2021}. 
Jouini et al.~\cite{jouiniReconstruction2013} proposed the Self-Organizing Map (SOM) network to complete Chl-a data under heavy cloud coverage by integrating SST and sea surface height (SSH). 
Artificial Neural Network~(ANN) is applied for the completion of SST data in the Mediterranean~\cite{pisoniArtificial2008}. 
Hankel Matrix Factorization (HMF) is introduced for data completion in the Global Navigation Satellite System (GNSS)~\cite{liuMissing2022}.
 
In sum, traditional machine learning-based methods are easy to set up, have the ability to model the dynamic, non-linear, and noisy features of ST ocean data, and also achieve good performance on short-term data.
However, they have a high computational cost, tend to overfit when applied to long-term data, and cannot well capture the complex ST dependencies in ocean data.

\subsubsection{Deep learning-based methods} 
To capture the complex ST dependencies, deep learning-based data completion methods have been introduced to improve the quality of sparse ST ocean data.
For example, the Neural Classification method is proposed for filling the cloud covering values in SST data \cite{jouiniReconstruction2013}. 
Jean-Marie~\cite{vientDataDriven2021} achieved the completion of SST data using a neural network~(NN) and proved that the NN is superior to the OI and EOF methods. 
Zhao et al.~\cite{zhao2023integrated} proposed an LSTM-based method to fill the missing values of wave height data.
Barth et al.~proposed the Data-Interpolating Convolutional Auto-Encoder (DINCAE) for the completion of Chl-a data~\cite{barthDINCAE2020} and SST data~\cite{hanApplication2020}. 
Generative Adversarial Imputation Network (GAIN)~\cite{pan2021tec} with different generation rules has been introduced for missing data completion, and can effectively learn the complex data distribution.
GCN-based models~\cite{lian2023gcnet} have been adopted to utilize an algebraic framework termed Graph-Tensor Singular Value Decompositions (GT-SVD) to extract hidden spatial information in ocean monitoring data.
Wang et al.~\cite{wang2022sta} combined attention and GCN~\cite{wang2022sta} to capture both short-term temporal dependence and dynamic spatial dependence in ST ocean data and achieved good performance in satellite data completion.
Moreover, the diffusion model has been applied in this field due to its superior generative ability. 
Tashiro et al.~\cite{tashiro2021csdi} proposed a score-based diffusion model for filling incomplete values and developed a self-supervised training method by using available observation values to train the model.

Deep learning-based data completion algorithms could achieve high accuracy in ST ocean data completion and can capture dynamic dependencies over time. However, they still have some weaknesses. First, deep learning-based data completion algorithms are of high complexity, and their training and tuning processes are time-consuming. Besides, the deep learning methods are treated as "black boxes" and have low interpretability.
Therefore, it is still worth exploring to reduce the time complexity of deep ST ocean data completion methods and enhance models' interpretability in the future.

\subsection{Data Fusion}\label{sec:datafusion}
For many STDM tasks, e.g., climate forecasting and typhoon tracking, multi-source ocean data collected by various producers, e.g., NOAA, NASA, and ECMWF, are required to obtain comprehensive knowledge.
These different datasets usually have diverse data collecting standards, equipment, and technologies, leading to different ST resolutions,  different ST coverage, and different data quality. 
For example, in-situ observations can provide precise information about the ocean at specific locations, but they have issues of sparsity, uneven distribution, and low resolution. Meanwhile, satellite data can provide continuous observation of the ocean over a wide area and for long periods, but is unable to obtain fine-grained information at the location level.
Therefore, how to combine the benefits of multi-source observation data to build high-quality ST ocean datasets is an important yet challenging problem.
Specifically, for STDM in ocean, data fusion is to integrate multiple data sources to obtain more comprehensive and consistent information about the ocean.
Existing methods for fusing multi-source ST ocean data can be roughly classified into two categories, i.e., homogeneous data fusion methods and heterogeneous data fusion methods.

\subsubsection{Homogeneous data fusion methods} 
Homogeneous data fusion aims to combine the datasets from the same source into a comprehensive dataset to improve the data quality.
At early ages, the most common way to fuse homogeneous data is to utilize their statistical characteristics~(e.g., local mean matching, regression analysis, and statistical region merging)~\cite{cressie1996change}.
However, it is difficult to fuse the observation records of two datasets from the same source with different spatial and temporal scales.
To solve this problem, Nguyen et al.~\cite{nguyen2012spatial} presented the spatial data fusion methodology based on the spatial-random-effects model, where the spatial distribution is learned based on the data of different temporal scales within the same region.
Kang et al.~\cite{ma2020spatio} proposed the Dynamic Fused Gaussian Process (DFGP) to combine a low-rank representation with a general covariance matrix with a dynamic-statistical approach to fuse the satellite data from different sensors.
Jung et al.~\cite{jung2022high} utilized the RF to fuse two satellite datasets to acquire high-quality SST data.
To extract fine-grained data information from ocean data, many researchers utilize deep learning methods to learn the data embedding to fusion the data for downstream tasks~(e.g., SST prediction).
Hou et al.~\cite{hou2022must} proposed an ST ocean data fusion model based on ConvLSTM to combine three different satellite datasets and achieve good SST prediction results.
Raizer et al.~\cite{raizer2013multisensor} proposed a digital wavelet transform method for combining the information in both spatial and temporal domains on two satellite datasets.

\subsubsection{Heterogeneous data fusion methods} 
Heterogeneous data fusion aims to combine data from different data sources, thus providing more comprehensive information than the dataset from a single source.
Due to the variety in form and content across spatial and temporal scales, it is much more difficult to fuse heterogeneous data than homogeneous data.
Wikle and Berliner~\cite{wikle2005combining} proposed a hierarchical Bayesian model to fuse the satellite data with reanalysis data by adjusting the data of different spatial resolutions with its areal-averaged data.
McMillan et al.~\cite{mcmillan2010combining} proposed an ST model that combines in-situ observations and the reanalysis model's gridded outputs by assuming that the data collected at the same location has the same data distribution.
Chen et al.~\cite{10.2112/SI79-024.1} provided a Bayesian inference model to fuse the in-situ data and satellite data for the accurate estimation of chlorophyll-a. 
Han et al.~\cite{han2021sea} proposed a CNN-based fusion algorithm to combine the satellite image data and time-series data and achieved good performance in downstream applications.
Wang et al.~\cite{wang2023fusion} provided an ODF-net to fuse in-situ observations, satellite observations, and reanalysis data with a self-training attention mechanism.

In conclusion, fusing multi-source data can enhance data consistency and quality, and improve the accuracy of STDM tasks for ocean science.
However, existing methods for fusing ST ocean data usually rely heavily on prior statistical knowledge of linear principles, normal distributions, and error covariances.
In addition, these methods are often tailored for specific datasets and tasks, and there are no unified data fusion frameworks that can fuse various ST ocean data sources. These limitations are still open issues that need further exploration.

\subsection{Data Transforming}
\begin{figure}[]
\begin{minipage}[t]{\textwidth}
\centering
\includegraphics[width=0.7\textwidth, height=0.35\textwidth]{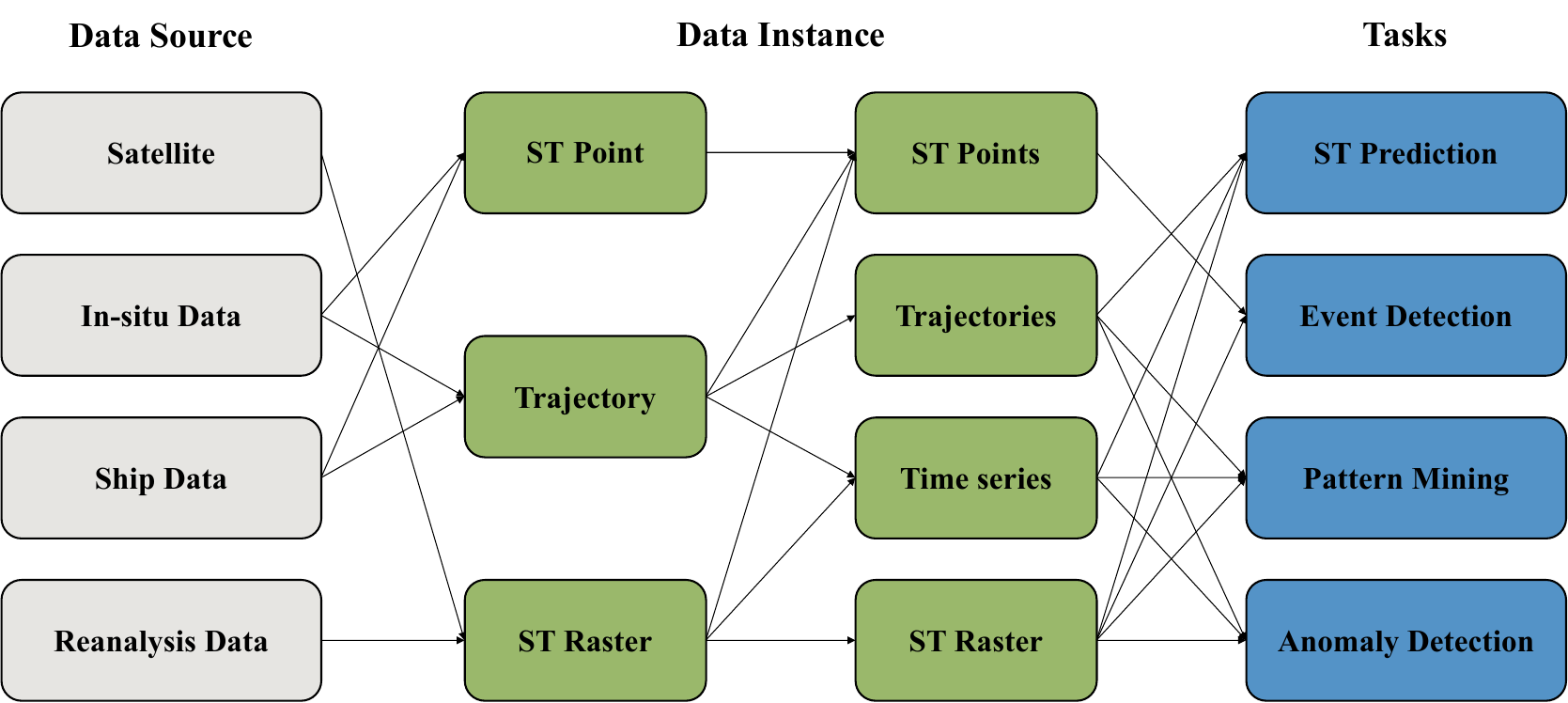}
  \caption{A mapping from the data sources to the data instances and the four types of ocean STDM tasks.}
    \label{Transforming}
\end{minipage}
\end{figure}

In practice, different ocean application scenarios correspond to different categories of STDM tasks and problem formulations, 
and different STDM tasks usually have different requirements for the formats of input data.
Therefore, the original data from multiple sources cannot be directly used for various STDM tasks.
An important process before putting data into STDM models is to transform the data to meet the specific requirements of the data mining task.
As shown in Fig.~\ref{Transforming}, we provide a clear mapping from different data sources to different data instances~(i.e., ST points, Trajectories, Time series, and ST raster) and four types of tasks~(cf. Section~\ref{sec:tasks-methodologies} for the details).
For example, an ST point refers to a tuple containing spatial and temporal information as well as ocean features~(e.g., temperature and chlorophyll-a), which can be obtained by in-situ sensors and ship sensors. 
Then, a series of ST points collected at different locations can be utilized for detecting various events~(e.g., El Ni$\mathrm{\tilde{n}}$o and typhoon).
A trajectory, usually collected by a sailing ship or moving sensor, consists of the continuous measurements of an ST feature over a set of moving reference points in space and time and can be utilized in ship anomaly detection, and pattern mining tasks.
ST raster data refers to the measurement of a continuous or discrete ST field recorded at fixed locations in space and at fixed points in time, which is usually collected by the satellite sensors and obtained from the outputs of reanalysis models.
ST raster data is an important data format for STDM tasks and also can be transformed into other data instances~(e.g., ST points and time series) for different tasks.

\begin{figure}[]
\begin{minipage}[t]{\textwidth}
\centering
\includegraphics[width=\textwidth, height=0.65\textwidth]{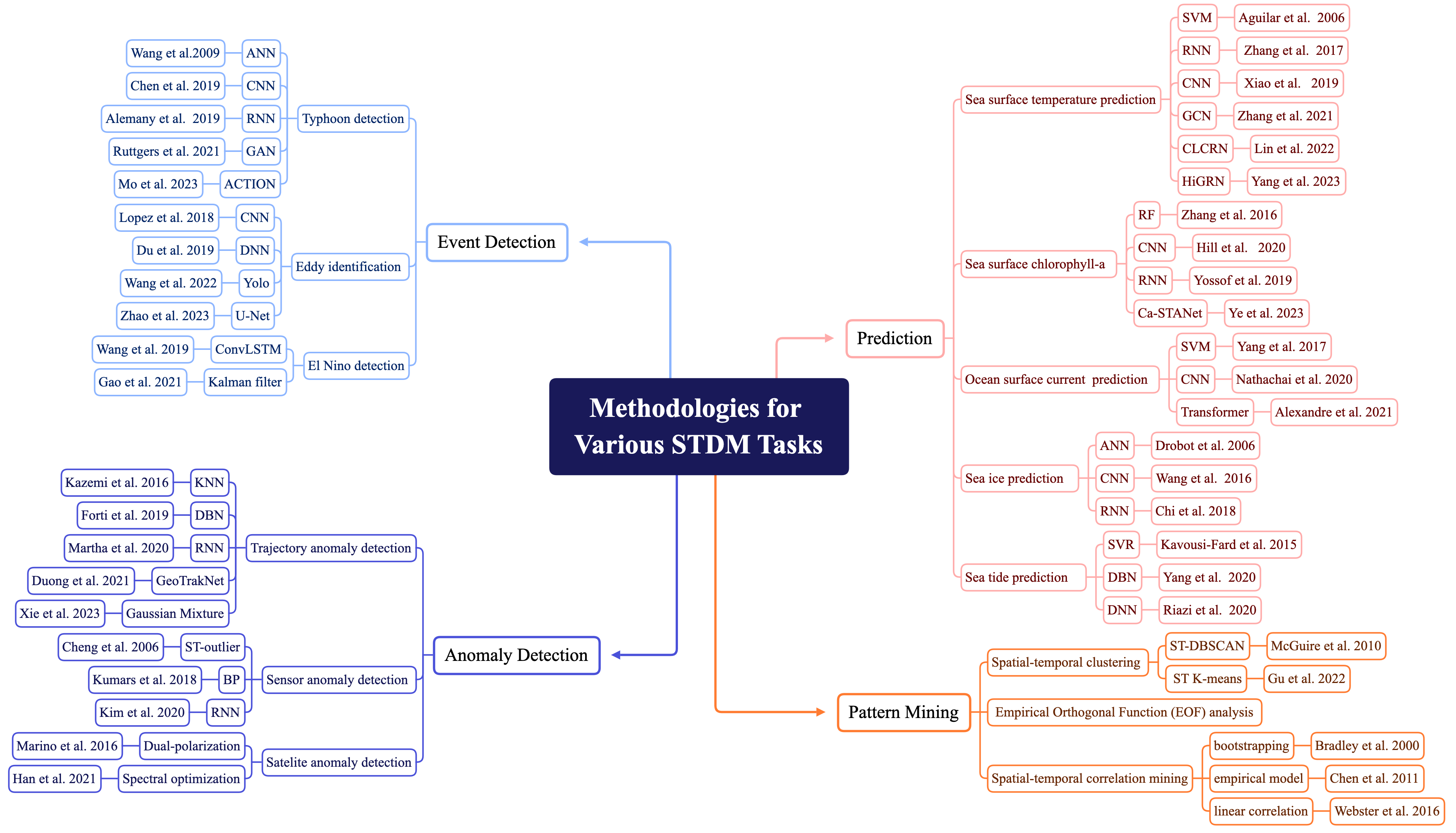}
  \caption{A taxonomy of methodologies of different ocean STDM tasks.}
    \label{taxonomy}
\end{minipage}
\end{figure}

\section{Tasks \& Methodologies }\label{sec:tasks-methodologies}
\begin{figure}[]
\begin{minipage}[t]{\textwidth}
\centering
\includegraphics[width=0.6\textwidth]{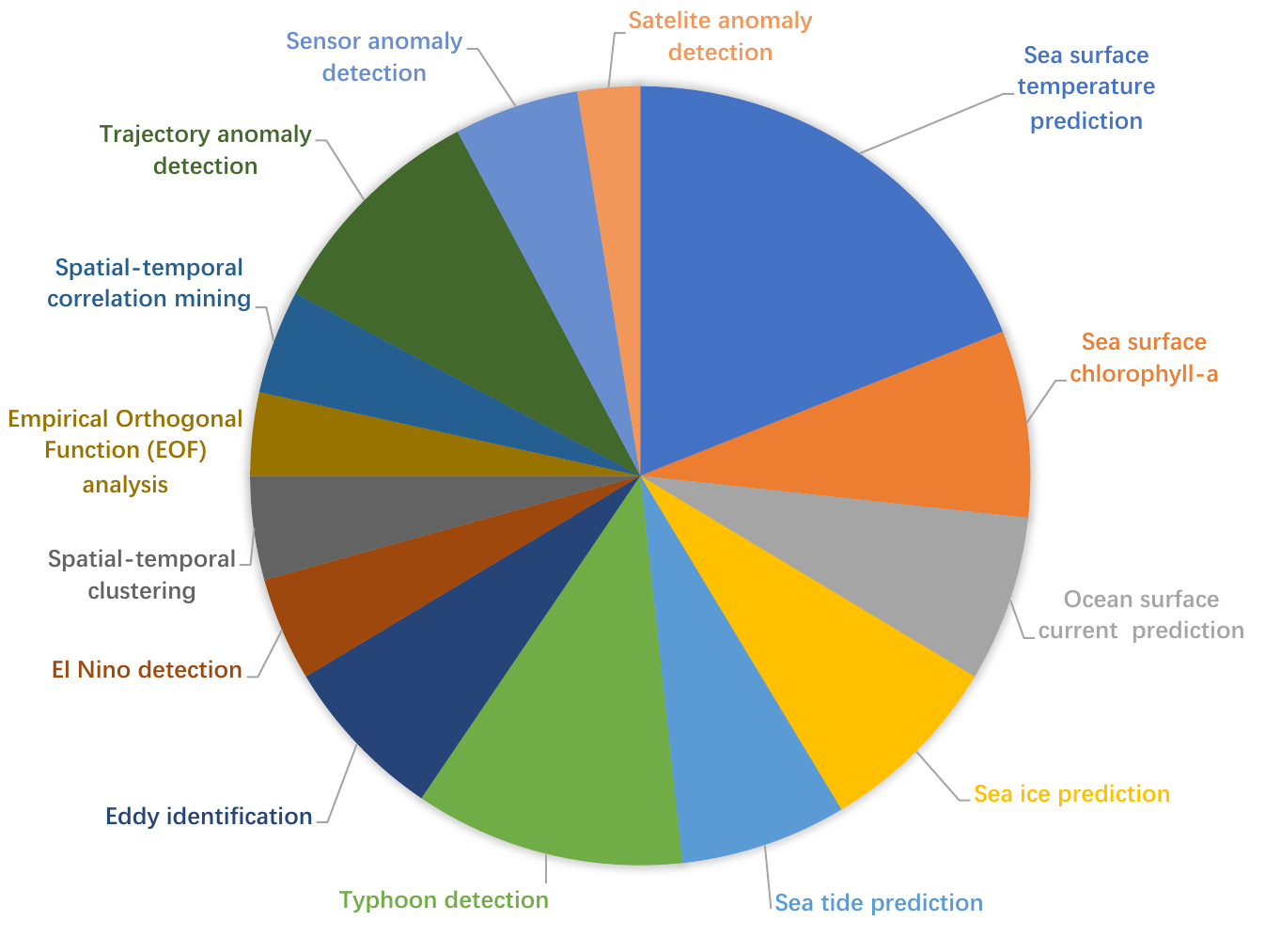}
  \caption{The distribution of publications of various ocean STDM tasks (data from the Web of Science.}
    \label{Distributions}
\end{minipage}
\end{figure}

As shown in Fig.~\ref{taxonomy}, to provide a clear overview of the STDM techniques for ocean science, we classify the STDM tasks
in ocean into four typical categories, i.e., prediction, event detection, pattern mining, and anomaly detection, based on the problem formulation.
And Fig.~\ref{Distributions} illustrates the distribution of published research papers on these tasks, among which SST prediction and typhoon detection are the most popular.
In addition, Table~\ref{tab-works} further presents the representative tasks for each category of STDM tasks.



\begin{table*}
\centering
\caption{Representative ocean STDM tasks.}
\label{tab-works}
\resizebox{\linewidth}{!}{
\begin{tabular}{ccc}
\hline
Task Categories                                               & Tasks                                  & Related studies \\ \hline
\multirow{4}{*}{Spatial-temporal Prediction}                    & Sea surface temperature prediction                   &  \makecell[c]{\cite{linGFDL},\cite{stortoCMCC},\cite{stortoEvaluation2016},\cite{BUONGIORNONARDELLI20131},\cite{Zivot2003}, \cite{ARIMA}, \cite{mar},\cite{SVR},\cite{linsPredictionSeaSurface2013}, \\ \cite{WU2006145}, \cite{LSTMSST},  \cite{CFCCLSTM},\cite{zhangMonthlyQuarterlySea2020},\cite{cnn1},\cite{ConvlstmSST},\cite{D2CL},\cite{MGCN},\cite{TGCN}, \\ \cite{haghbinApplications2021}, \cite{10.2112/SI79-024.1}, \cite{koner2019daytime},\cite{hussein2021spatiotemporal}}.          \\
                                                                & Sea surface chlorophyll-a  prediction                   &    \cite{1621081},\cite{LIU2010681},\cite{10.2166/hydro.2002.0013}, \cite{su8101060},\cite{9113293},\cite{ijerph18147650},\cite{lee2018improved},\cite{wang2020unsteady},\cite{hussein2021spatiotemporal}.        \\
                                                                & Ocean surface current  prediction                      & \cite{frolov2012improved},\cite{barrick2012short},\cite{SARKAR2018221},\cite{guozhen2017tidal},\cite{REMYA201262},\cite{https://doi.org/10.1029/2018JC014471},\cite{immas2021real},\cite{8864215}.              \\
                                                                & Sea ice prediction                                     &  \cite{zakhvatkina2019satellite},\cite{lindsay2008seasonal},\cite{wang2016sea},\cite{drobot2006long},\cite{wang2017ice},\cite{petrou2017prediction},\cite{chi2017prediction}, \cite{liu2021daily},\cite{liu2021daily}.             \\
\multicolumn{1}{l}{}                                            & Sea tide prediction                                    &     \cite{LEE20021003},\cite{sarkar2016machine},\cite{7882673,kavousi2016modeling},\cite{granata2021artificial},\cite{okwuashi2017tide},\cite{yang2021sea},\cite{riazi2020accurate}.          \\ \hline
\multirow{3}{*}{Spatial-temporal Event Detection} & Typhoon detection                            &     \makecell[c]{ \cite{ali2007predicting},\cite{wang2011back},\cite{kim2011pattern},\cite{yang2017trajectory},\cite{chu2012integration},\cite{loridan2017machine}, \cite{kim2019deep},\cite{alemany2019predicting},\\ \cite{chen2019hybrid}, \cite{koner2019daytime},\cite{baik2000neural},\cite{ruttgers2018typhoon},\cite{giffard2018fused}. }        \\
                                                                & Eddy identification                          &    \cite{WEISS1991273},\cite{CHELTON2011167},\cite{rs13234905},\cite{13004},\cite{article11123},\cite{8518411},\cite{8613216},\cite{DU201989}.           \\
                                                                & El Ni${\mathrm{\tilde{n}}}$o detection \& prediction                         &    \cite{gao2021parameter},\cite{cane1986experimental},\cite{tang2018progress},\cite{wang2021hybrid},\cite{kirtman2003cola}.                     \\ \hline
\multirow{3}{*}{Spatial-temporal Pattern Mining}                 & Spatial-temporal clustering                   &     
\cite{hoffman2005using},\cite{white2005global},\cite{mcguire2010spatiotemporal},\cite{DBLP:journals/sensors/GuQZYZ22},\cite{9068243},\cite{10.2112/SI79-024.1}          \\
                                                                & Empirical Orthogonal Function (EOF) analysis &    \cite{cressie2015statistics},\cite{mestas1999rotated},\cite{rs13020265},\cite{8684281}.           \\                                                                
                                                                & Spatial-temporal correlation mining            &   \cite{article22312r},\cite{doi:10.1126/science.1116448},\cite{doi:10.1126/science.1209472},\cite{doi:10.1126/science.1123560},\cite{doi:10.1126/science.253.5018.390}.  
                                                               \\ \hline
\multirow{3}{*}{Spatial-temporal Anomaly Detection}              & Trajectory anomaly detection                 &      \makecell[c]{\cite{soleimani2015anomaly},\cite{kazemi2013open},\cite{wolsing2022anomaly},\cite{han2021modeling},\cite{ferreira2022novel},\cite{anneken2015evaluation},\cite{rong2019ship},\cite{rong2020data},\cite{forti2021bayesian}, \\\cite{soleimani2015anomaly},\cite{nguyen2021geotracknet}.    }    \\
                                                                & Sensor anomaly detection                     &\cite{MahmoodiGhassemi+2018+44+50},\cite{barua2007parallel},\cite{DBLP:conf/kdd/WuLC08},\cite{anbarouglu2009spatio},\cite{kim2021outlier},\cite{cheng2006multiscale}.
\\
                                                                & Satelite anomaly detection                   &   
\cite{marino2016depolarization},\cite{han2021spectral},\cite{fahn2019abnormal}.            \\ \hline 
\end{tabular}
}
\end{table*}

\subsection{Spatial-temporal Prediction}
Spatial-temporal prediction is an important STDM task for ocean science and can predict the future change of many ocean factors~(e.g., temperature, chlorophyll-a, and ice concentration).
ST prediction aims at understanding the regularity of past observations and predicting future observations, enabling related decision-making and applications. 

Given the time series ${\mathcal{X}} = \{X_{:0}, X_{:1},\cdots, X_{:t}\}$, we have  $X_{:t} = \{x_{1,t},$
$ x_{2,t},\cdots,x_{N,t}\} \in \mathbb{R}^{N\times d}$ that records $d$-dimension data at $N$ locations at time $t$, where $d$ represents the dimension of observed factors. Then, ST prediction can be formulated as seeking a function \bm{$\mathcal{F}$} to predict these features in upcoming  \bm{$\tau$} time steps based on the historical data of the last $T$ time steps, i.e.,
\begin{equation}
  \{\hat{X}_{:t+1}, \hat{X}_{:t+2},\cdots, \hat{X}_{:t+\tau}\} = \bm {\mathcal{F}_{\theta}} \{X_{:t-T+1}, X_{:t-T},\cdots, X_{:t}\}
 \end{equation}
where \bm{$\theta$} denotes all the learnable parameters in the prediction function  \bm{$\mathcal{F}$}. 


The key challenge in ST prediction is that ocean factors have complex spatial and temporal dependencies. 
For example, events occurring in a region may affect other regions far away (spatially and temporally). 
For ST prediction in ocean, we overview five representative tasks, i.e., sea surface temperature prediction, sea surface chlorophyll-a prediction, ocean surface current prediction, sea surface chlorophyll-a prediction, and sea tide prediction, which are of high importance for various ocean applications. 

\subsubsection{Sea surface temperature prediction}
Sea surface temperature~(SST) is one critical parameter for monitoring global climate change, and accurate SST prediction is important to weather forecasting, disaster warning, and ocean environment protection.
The major challenge in predicting SST is that SST has dynamic patterns changing over time, and tends to have long-term temporal dependencies and complex spatial dependencies.
SST prediction has been studied for decades by many researchers and existing methods for SST prediction can be roughly grouped into three categories, i,e., physical models~\cite{linGFDL, stortoCMCC, stortoEvaluation2016, BUONGIORNONARDELLI20131}, temporal methods~\cite{Zivot2003, ARIMA, mar, SVR, LSTMSST}, and spatial-temporal methods~\cite{lin2022conditional, CFCCLSTM, D2CL, MGCN}. 

The main idea of physical models is to combine the laws of physics, e.g., Newton's laws of motion, the law of conservation of energy, and the seawater equation of state, to predict SST.
For instance, the Global Forecast System (GFS)~\cite{linGFDL} conducts SST prediction by combining the Navier-Stokes equation, solar radiation function, and ocean latent heat circulation equation to simulate the changes of SST.
Although physical models are widely used in the past decades, they require a good understanding of the underlying mechanism of SST to choose the determining factors of physical functions.
However, the mechanism is dynamic and complicated, which makes it difficult to use only the factors of explicit equations to capture SST patterns.

Temporal methods, as a widely used type of SST prediction methods, formulate SST prediction as a time series prediction problem that can be solved by various temporal models, e.g., vector autoregressive models~\cite{Zivot2003}, autoregressive integrated moving average (ARIMA)~\cite{ARIMA},  hidden Markov models~(HMM)~\cite{mar}, support vector machines~(SVM) model~\cite{SVR}, and some temporal deep learning methods~\cite{lin2022conditional, LSTMSST, CFCCLSTM}.
For example, Xue et al.~\cite{mar} proposed a seasonally varying HMM method to construct a multivariate space of the observed SST and achieved good prediction performance.
Aguilar et al.~\cite{WU2006145} introduced warm water volume as an additional feature to enhance SST prediction with an SVM model. 
Generally speaking, these are linear models that use a window of past information to predict the SST records in the future. 
Although these methods can predict the trend of SST to a certain extent, they ignore the long-term dependencies, leading to low overall prediction accuracy.
Zhang et al.~\cite{LSTMSST} employed a fully connected LSTM to predict future SST in Bohai, where the LSTM structure models SST sequences and the fully connected structure produces the prediction results.
These temporal methods mainly focus on extracting the temporal features of SST and cannot well capture the spatial correlations among the SST time series.

\begin{figure*}[]
  \centering
\includegraphics[width=0.9\textwidth]{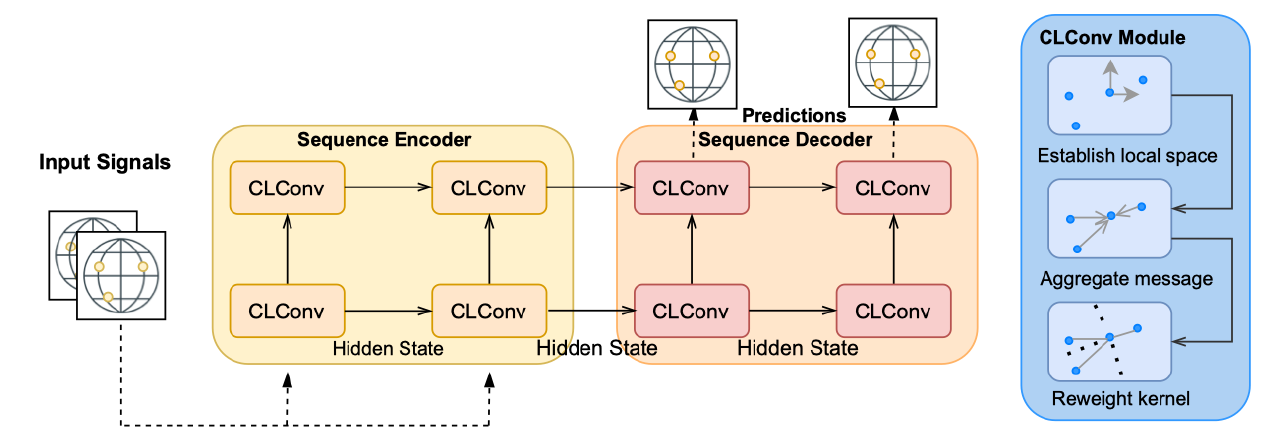}
  \caption{The framework of the CLCRN model, which is implemented based on the Sequence to Sequence architecture with the encoder fed with previously observed data and the decoder generating the predictions~\cite{lin2022conditional}. }
    \label{stp}
\end{figure*}

Spatial-temporal prediction methods combine typical deep learning methods such as CNN~\cite{cnn1}, RNN~\cite{elman1990finding}, and GCN~\cite{GCN} to capture the ST dependencies in ocean data.
Xiao et al.~\cite{ConvlstmSST} employed the ConvLSTM model for SST prediction by using CNN to extract the spatial information of gridded SST and LSTM to capture the temporal dependencies of SST.
Hou et al.~\cite{D2CL} proposed a dilated convolutional model to obtain long-term dependencies of SST series and the model achieves better prediction performance than ConvLSTM.
To learn the complicated non-adjacent connections among SST series, researchers also model the spatial correlations of SST time series with GCN.
As shown in Fig.~\ref{stp}, CLCRN~\cite{lin2022conditional} utilizes a GCN-based method to establish local spaces for modeling the dependencies of the spatial neighbors and predicting future SST.
Zhang et al.~\cite{MGCN} developed the memory GCN to build a distance-based adjacency matrix to represent the spatial correlations of SST series at different locations.
Yang et al.~\cite{10.1145/3597937} proposed the HiGRN to capture the hierarchical relationship between regions and achieved good performance.
Spatial-temporal prediction methods have been wide-used to capture the ST dependencies in SST and achieve good performance.
However, most of them focus on capturing the static dependencies in SST, ignoring the dynamic ST dependencies and the combination with the physic models, which is still worth exploring in the future.

\subsubsection{Sea surface chlorophyll-a prediction}
Predicting the sea surface chlorophyll-a (Chl-a) concentration could facilitate the monitoring and early warning of harmful algal bloom events which have serious impacts on marine fisheries. 
For Chl-a prediction, the major difficulty is to model the seasonality and the related factors affecting Chl-a.
Vollenweider~\cite{vollenweider1975input} et al. proposed a simple first-order equation to predict Chl-a by adjusting the parameters of the equation. ARIMA is one of the most common predictive models for predicting the Chl-a and has been applied to predict the distribution of Chl-a in the Japan Sea based on satellite data~\cite{1621081}.  
Although these models are able to predict the linear dependencies in Chl-a, they do not consider other factors affecting Chl-a.  
Liu et al.~\cite{LIU2010681} used the absolute principal components score~(APCS) method to simulate the effects of multiple chemical variables on Chl-a in Lake Qilu. 
However, these statistical methods are of low prediction accuracy.

In recent years, various machine learning methods~(e.g., neural networks,  SVM, decision trees~(DT), RF, feed-forward neural network~(FFNN), and regression models) are widely used for Chl-a prediction~\cite{su8101060}.
For example, an FFNN model~\cite{10.2166/hydro.2002.0013} was used to forecast the Chl-a concentration seven days in advance in the Lake Kasumigaura of Japan and obtained better performance than the statistical methods.
These models can capture the complex relationship by training the parameters with past observations and achieving good performance. 
To further explore the ST dependencies between different ocean locations,
Hill et al.~\cite{9113293} used a CNN to predict Chl-a concentrations within the Gulf of Mexico region and achieved better performance than traditional machine learning methods.
Yossof et al. \cite{ijerph18147650} adapted an LSTM model and a CNN model to predict the harmful algal blooms on the western coast of Sabah. 

To well learn the seasonality and tendency of Chl-a for long-term prediction, we need to capture the dynamic dependencies of Chl-a over time.
Liu et al.~\cite{na2022long} combined the CNN-LSTM model with ST trend analyses, and increased the prediction accuracy of Chl-a in future 3 years.
Wang and Xu~\cite{wang2020unsteady} used a dual-stage attention-based RNN (DA-RNN) to predict Chl-a concentration. The results show that the LSTM model outperforms the CNN model in terms of prediction accuracy. 
Ye et al.~\cite{10083105} proposed the Spatiotemporal attention network predict the Chl-a with gap-filled Chl-a data and achieve high long-term accuracy.
Although they are many existing methods adapting machine learning algorithms to predict Chl-a, they seldom model the relationship between Chl-a and other ocean factors~(e.g., temperature and PH).
Moreover, due to the difference in hydrology, climate, geochemistry, and biological characteristics of different ocean regions, existing Chl-a prediction methods may not have the transferability to be suitable for all ocean regions. In this case, how to design a general Chl-a prediction model is still an ongoing research problem.

\subsubsection{Ocean surface current prediction}
Ocean surface current refers to the large-scale, regular, and stable flows of water on the surface of ocean in a certain direction throughout the year.
Accurate prediction of ocean currents can not only assist shipping, fishery, and tourism to produce economic benefits but also play an important role in many marine tasks such as vessel path planning and Autonomous Underwater Vehicles~(AUS) controlling.
The major challenge in ocean surface current prediction is to model the dynamics of slow and large-scale currents.
Earlier studies on ocean current prediction mostly used numerical modeling methods such as physical models of ocean circulation and linear statistical models.
For example, the physical methods calculate currents from the sea surface height with physical equations~\cite{le2001ocean, 10.3389/fmars.2021.672477}.
As for statistical models, Frolov et al.~\cite{frolov2012improved} developed an ocean surface current prediction model based on the linear autoregressive technique. 
Barrick et al.~\cite{barrick2012short} presented a short-term statistical system that only requires a few hours of previous data to predict future ocean surface currents in Northern Norway. 
Nevertheless, these physical and statistical methods require substantial processing expenses and predetermined parameters that must be manually given by domain experts.

With the development of machine learning, non-parametric prediction methods are gradually used for ocean current prediction.
Many machine learning algorithms, e.g., Gaussian Processes, SVM, Genetic Algorithm, and Artificial Neural Network~(ANN) have been utilized to improve the prediction accuracy of currents~\cite{SARKAR2018221,guozhen2017tidal, REMYA201262}. 
For example, Sarkar et al.~\cite{https://doi.org/10.1029/2018JC014471} proposed an ANN method with fully connected layers and recurrent LSTM layers to predict the tidal currents in a given region and achieved good results. 
Alexandre et al.~\cite{immas2021real} proposed a Transformer model with LSTM to capture the ST dependencies using real-time in-situ data to predict ocean currents. 
Nathachai et al.~\cite{8864215} proposed an ST model that takes advantage of noticeable domain characteristics by using a CNN to capture the spatial property and the GRU to capture the temporal property.

Although existing studies can already capture the ST information in ocean surface current data and achieve good prediction performance, the large-scale spatial dependencies and the correlations with other related factors~(e.g., wind flow and the ocean tides) are still not well considered, which affect the prediction accuracy of ocean surface currents.



\subsubsection{Sea ice prediction} Sea ice is simply frozen ocean water. 
Sea ice prediction is a crucial task to monitor and prevent global warming and also plays an important role in coastal port traffic and vessel path planning.
For the study of sea ice,  data from satellites are often the main or even the only data source because it is difficult to obtain in-situ sea ice observation data in extreme weather.
According to~\cite{mu2023icetft}, the ever-increasing melting speed has made it more difficult to make accurate sea ice predictions. 
Consequently, how to utilize the limited data to predict future sea ice data is a challenging issue.

Currently, the majority of the studies on sea ice prediction are data-driven methods~\cite{lindsay2008seasonal}.
Traditional machine learning methods, e.g., BP network, ANN, and linear regression~\cite{drobot2006long}, are widely used for sea ice prediction.
For example. Li et al.~\cite{li2021machine} provided a sea ice-intensive degree-day prediction method that combines kernel Granger Causality analysis (KGC) and SVM to predict the sea ice.
However, these machine learning methods ignore the ST dependencies in sea ice data, which leads to low prediction accuracy.
To further capture the spatial dependencies, Wang et al.~\cite{wang2016sea} proposed a CNN-based method to estimate the sea ice in the melting seasons by using image data from satellites. 
Wang et al.~\cite{wang2017ice} provided the fully convolutional neural networks (FCNN) model to predict sea ice along the east coast of Canada and the model achieves high prediction accuracy. 
To capture the temporal dependencies in sea ice data, Petrou et al.~\cite{petrou2017prediction} used ConvLSTM to learn the temporal information and obtain good performance in short-term sea ice prediction. 
Chi et al.~\cite{chi2017prediction} proposed a monthly sea ice prediction model using MLP and LSTM, and achieved good performance on the arctic sea ice concentration data. 
To identify the impactful factors on sea ice, Mu et al.~\cite{mu2023icetft} proposed an interpretable long-term prediction model (IceTFT) which fuses multiple atmospheric and oceanic variables as input to train the model, and utilizes the temporal fusion transformer and the multi-head attention mechanism to automatically adjust the weights of predictors and filter spuriously correlated variables.

Most of the current studies on sea ice prediction are purely data-driven methods~\cite{liu2021daily}.
However, it has been proven that combining physical knowledge with machine learning is an effective way to achieve accurate sea ice prediction. Currently, how to combine the physical mechanisms of sea ice with data-driven methods remains a challenging issue for achieving accurate sea ice prediction.

\subsubsection{Sea tide prediction}
Tides are periodic fluctuations of the ocean surface and are caused by the gravitational pull of the sun and moon. Tidal energy generated by the fluctuation of tidal water level is a vital renewable energy source~\cite{LEE20021003}.
For a long time, tide prediction provides safety guarantees for port operations, shipping traffic, and coastal protection, and also provides important support for offshore aquaculture and the prevention of natural disasters.

In the early ages, the most commonly used technique to predict sea tides is harmonic analysis which forecasts the tidal variations with historical observations.
However, harmonic analysis suffers from the fact that the parameters are numerous and some of which have a very long return period, thus costing a lot of time.
Currently, the widely-used methods for tidal prediction are also based on machine learning. 
For example, Sarkar et al.~\cite{sarkar2016machine} used the BP algorithm, cascade correlation algorithm, and conjugate gradient algorithm to predict the tides continuously. 
Kavousi-Fard et al.~\cite{kavousi2016modeling} proposed a probabilistic method that uses Lower Upper bound Estimation (LUBE) to model the uncertainties of tidal current prediction, and achieves good prediction performance in the Bay of Fundy, NS, Canada.
Granata et al.~\cite{granata2021artificial,okwuashi2017tide} used tree regression, RF, and multilayer perceptron to predict the tide level of Venice based on satellite data. 

Due to the high sparsity and large spatial resolution of satellite data, it is difficult to build a high-quality unified sea tide prediction model. 
Meanwhile, the geology and geomorphology of the local environment are also crucial for predicting the sea tide.
However, these methods above do not consider multiple source data and external factors, resulting in low prediction accuracy. 
To address this issue, some researchers integrate multi-source sea tide data and external factors to predict tides. 
For example, Yang et al.~\cite{yang2021sea} integrated satellite data and tide gauge data to estimate tidal sea level using a Deep Belief Network~(DBN).
Riazi et al.~\cite{riazi2020accurate} built a DNN to achieve tidal level prediction by using tidal generating force, and geological and biological factors as inputs.
However, the deep ST correlations in sea tide data are still not fully utilized by existing methods to improve the prediction accuracy for sea tides.


In a word, with the development of various STDM methods, ST prediction has gained a lot of attention and boosted the prediction accuracy of various ocean factors as discussed above.  
Further uncovering the hidden and deep ST correlations in ST ocean data and fusing them into the prediction models are still an open problem for ST prediction in ocean.

\subsection{Spatial-temporal Event Detection}
In ocean science, an ST event usually means that there are significant and persistent changes happening at a particular time and location, such as typhoons, eddies, and other extreme ocean phenomena.
Monitoring and detecting these events can provide early warning to prevent severe damage to personal safety and social safety.
In recent years, extreme ocean events become more and more frequent under the circumstance of global warming.
Therefore, accurate detection and automatic identification of ST events in ocean are of significance and attract a variety of researchers. 
Specifically, in this section, we give a brief overview of the detection of typical ST ocean events, i.e., typhoons, eddy, and El Ni$\mathrm{\tilde{n}}$o event.

\subsubsection{Typhoon detection}
A typhoon refers to a large storm system that has a circular or spiral system of violent winds and typically has hundreds of kilometers in diameter. Typhoon often occurs in the tropical ocean and is usually accompanied by strong winds, heavy rain, and storm surge.
Typhoon detection is of high importance to provide early warnings for disaster management in coastal areas.
Typhoons usually move rapidly and are affected by multiple environmental parameters, e.g., SST, cloud cover, atmospheric pressure, and wind speed. 
To detect a typhoon, it is crucial to capture the extreme changes of these factors at a particular time and location. 
Typhoon detection has been studied for decades and existing methods can be roughly grouped into two categories, i.e., statistical-based methods and image-based methods, according to the type of data.

For the statistical-based methods, typhoon-related atmospheric and oceanic factors~(e.g., SST and wind speed) and tide-gauge data are collected for the detection of typhoons.
For example, Ali et al.~\cite{ali2007predicting} used ANNs and BP methods to achieve the 24-hour track prediction of cyclones in the Indian Ocean and the Northwest Pacific Ocean. 
Alemany et al.~\cite{alemany2019predicting} established a fully connected RNN by using typhoon center location, wind speed, and central pressure at 6-hour intervals from NOAA to reduce truncation error in classical numerical prediction methods. 
Sophie et al.~\cite{giffard2018fused} combined the NOAA unifying tropical cyclone data with the ERA reanalysis wind field data to construct a fusion neural network for typhoon path prediction.
To predict the typhoon formation process and intensity, Chen et al.~\cite{chen2019hybrid} combined the CNN and LSTM to learn the spatial relationship and temporal relationship, respectively, among typhoon-related atmospheric and oceanic variables.
Moreover, many researchers formulate typhoon detection as an ST clustering problem, which can not only discover the typhoon but also localize central typhoon cyclones. 
Kim et al.~\cite{kim2011pattern} evaluated the typhoon detection performance of three different clustering methods~(i.e., K-means clustering, Fuzzy C-means clustering (FCM), and hierarchical clustering), and the results show that FCM is better in clustering efficiency and fitting degree. 
Yang et al.~\cite{yang2017trajectory} used the mixed regression clustering algorithm and the mass moment of the typhoon cyclone trajectory method to identify the typhoon tracks in the South China Sea.

With the development of computer vision and the increasing amount of satellite image data, CNN-based methods are widely used to discover typhoons.
Kovordanyi et al.~\cite{koner2019daytime} established a multi-layer ANN to predict the movement of cyclones based on the AVHRR images from NOAA. 
Ruttgers et al.~\cite{ruttgers2018typhoon} proposed the GAN-based detection method and found that combining the reanalysis data of physical models and satellite image data can improve detection performance. 
Kim et al.~\cite{kim2019deep} proposed a ConvLSTM-based ST model to ascertain the tracks of typhoons and predict their future positions. 
Considering that typhoons are relatively rare weather phenomena and the proportion of positive and negative samples could be seriously unbalanced, Zhao et al.~\cite{zhao2020real} proposed a CNN-based detection method with data enhancement technology, such as shearing, rotation, and gray-level transformation, to generate new samples for better typhoon events detection.

The state-of-the-art typhoon detection methods can already achieve quite good detection performance. However, achieving real-time typhoon detection is still a challenging problem due to the high complexity of detection methods and the complex dynamics in the development of typhoons. 

\subsubsection{Ocean eddy identification}

An eddy is a circular current of water.
Ocean eddies play an important role in ocean energy transfer and nutrient distribution. 
Accurately identifying the variation of ocean eddies is the key to understanding the oceanic flow and ocean circulation system~\cite{WEISS1991273}.
It is challenging to autonomously detect eddies since they are not static objects, but a distortion evolving through a continuous field.
Images obtained from satellite sensors have a large spatial resolution and a wide swath of observation, which makes them effective sources for gaining comprehensive and detailed information to detect ocean eddies.
Existing methods for ocean eddy identification can be roughly grouped into two categories, i.e., physical-based methods and data-driven methods.

The physical-based methods describe and detect the eddies with their physical information.
The Okubo-Weiss method~\cite{WEISS1991273} is the most popular statical and physical procedure to discover the patterns of eddies, and it also has good interpretability.
Chelton et al.~\cite{CHELTON2011167} tracked and investigated the eddies' variability globally using sixteen years of unified sea surface height data. 
Yang et al.~\cite{13004} proposed a logarithmic spiral edge fitting-based method to extract eddy information such as the center position, diameter, and edge size, from satellite images.  
These methods are dependent on the physical parameters, geometrics, and handcrafted features related to eddies. However, eddies are highly dynamic and the underlying physical processes are too complex to be described using the physical models with predetermined parameters. 

\begin{figure*}[]
  \centering
\includegraphics[width=\textwidth]{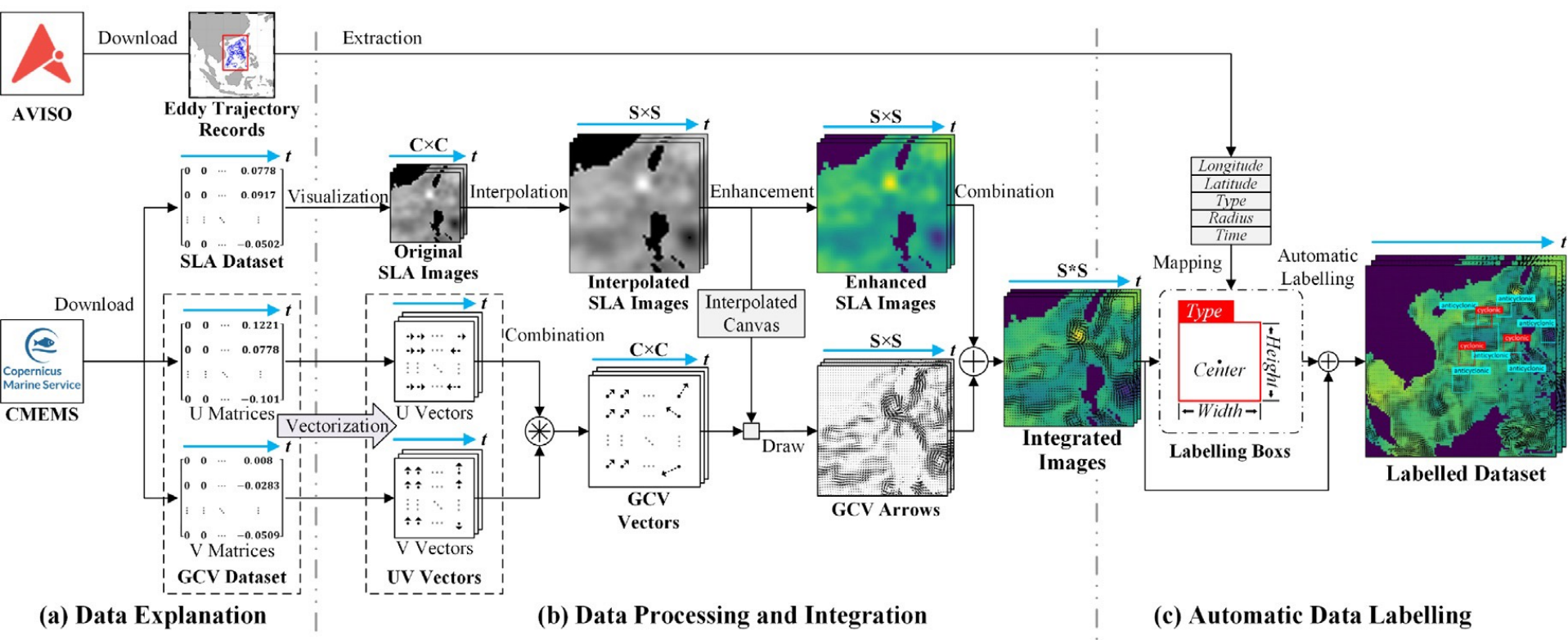}
  \caption{The framework of data-attention-YOLO for eddy identification. Here, (a) the two input datasets, (b) the data processing and integration process, and (c) automatic data labeling that is designed to transform the original eddy data to the labeled integrated data~\cite{wang2022data}.}
    \label{eddy}
\end{figure*}

Instead of extracting factors based on expert knowledge, data-driven models can learn high-level eddy features automatically from data.
For example, Du et al.~\cite{DU201989} utilized a spatial pyramid model and an SVM classifier method to recognize the eddies in satellite image data.
As shown in Fig.~\ref{eddy}, Wang et al.~\cite{wang2022data} proposed the object detection method using the Data-Attention-Yolo to process the satellite data and achieved high performance on eddy identification.
And Zhao et al.~\cite{10041953} developed the pyramid split attention (PSA) eddy detection U-Net architecture (PSA-EDUNet) that targets oceanic eddy identification from ocean remote sensing imagery and gain good result.
These data-driven methods utilizing satellite data are proven to be effective for eddy detection.
However, due to the limited resolution of satellite data, existing methods can only discover eddies with a large diameter but lack the ability to detect eddies with a diameter smaller than 50km.

\subsubsection{El Ni${\tilde{n}}$o detection}
El Ni$\mathrm{\tilde{n}}$o is an irregularly periodic variation in winds and SST over the tropical Eastern Pacific Ocean, and it can cause extreme weather events, e.g., floods and droughts, around the world. 
An El Ni$\mathrm{\tilde{n}}$o event can be identified if the absolute value of the 3-month mean of SST in the Ni$\mathrm{\tilde{n}}$o 3.4 region~($ 5^\circ$S-$ 5^\circ$N, $ 170^\circ$E - $ 120^\circ$W) reaches or exceeds $ 0.5^\circ C $ and lasts for at least 5 months.
Since the 1970s, there are many researchers working on El Ni$\mathrm{\tilde{n}}$o detection and various models have been proposed to identify the presence of a shift into El Ni$\mathrm{\tilde{n}}$o events in advance and predict the event duration and strength to allow early preparation for the related disasters. 
In general, El Ni$\mathrm{\tilde{n}}$o event has mutual influence with many other regions across the world.
Therefore, the biggest challenge for El Ni$\mathrm{\tilde{n}}$o detection is to capture the complex ST dependencies of the SST in different regions within a large spatial coverage. 

Most early El Ni$\mathrm{\tilde{n}}$o detection models use physical laws and mathematical equations to deduce the results from the complicated interaction between the atmosphere and ocean~\cite{kirtman2003cola}.
However, these methods need domain knowledge and require a lot of factor setting of functions.
To address this issue, many researchers use data mining methods for El Ni$\mathrm{\tilde{n}}$o detection, which do not involve physical equations and achieve better performance than the physical type~\cite{tang2018progress, cane1986experimental}.
For example. Cane et al.~\cite{cane1986experimental} proposed a linear dynamical model to identify the El Ni$\mathrm{\tilde{n}}$o event one year ahead.

Recently, with sufficient observational data and increasing computational power, deep learning methods are introduced to achieve long-term~(e.g., more than one year) prediction and detection of El Ni$\mathrm{\tilde{n}}$o events.
For instance, Wang et al.~\cite{wang2021hybrid} combined the ConvLSTM Encoder-Decoder model with the Empirical Mode Decomposition (EMD) technique to achieve El Ni$\mathrm{\tilde{n}}$o prediction three years in advance. 
Gao et al.~\cite{gao2021parameter}  proposed a new inflation scheme based on a local ensemble transform Kalman filter to improve the long-term prediction of El Ni$\mathrm{\tilde{n}}$o events.
Though these methods can provide valuable suggestions on whether El Ni$\mathrm{\tilde{n}}$o events will occur, it is still difficult to detect the onset of an El Ni$\mathrm{\tilde{n}}$o event because a series of reactions together lead to the El Ni$\mathrm{\tilde{n}}$o events.
Moreover, the existing detection models do not have good interpretability, which leads to poor confidence in the results.



\subsection{Spatial-temporal Pattern Mining}
ST patterns mining is a fundamental task of STDM to discover the hidden, potentially useful, and unknown associations and correlations among ST ocean data.
Classifying and identifying specific patterns or key features in ST ocean data are of great importance for various applications such as finding circulation regimes, discovering teleconnection patterns between weather phenomena at widely separated locations, identifying extreme-causing weather patterns, and evaluating the effects of climate changes.
In this section, we will review some notable ST pattern mining approaches, i.e., clustering, EOF analysis, and correlation mining, in ocean.

\subsubsection{Spatial-temporal clustering}
Clustering is one key data analysis task in data mining, which aims to group the input data samples and put the samples of high similarity into one class. 
In general, each data sample to be classified is defined as an input vector, and the clustering result is determined by calculating the similarity between the input vector and the clustered vectors. 
If the similarity reaches a certain value, the input vector is divided into the corresponding classes.  
Compared with linear regression analysis, clustering methods (e.g., hierarchical clustering, K-means clustering, and self-organizing maps clustering) are able to reveal unknown and hidden patterns without prior knowledge from domain experts.

At an early age, hierarchical clustering methods are utilized to discover the spatial patterns of ocean circulation in the northern hemisphere~\cite{mo1988cluster, cheng1993cluster}.
Hoffman et al.~\cite{hoffman2005using} developed a quantitative statistical clustering technique termed Multivariate Spatio-Temporal Clustering (MSTC) to identify the regions with similar environmental characteristics, e.g., the dry high-altitude weather of North American and Asian winters, warming weather in Antarctica and western Greenland.
White et al.~\cite{white2005global} utilized the K-means method to generate the climate and vegetation clusters which were used to infer phenological responses to climate changes. 
Gu et al.~\cite{DBLP:journals/sensors/GuQZYZ22}  provided a hybrid estimation method that combines the k-means clustering algorithm and an ANN model to reveal the seasonal variation patterns of SST and salinity in the Indian Ocean.
Moreover, Rafael et al.~\cite{OceanCirculation} applied the dual SOM method to analyze the ocean salinity and currents data from 1993 to 2012 to discover the ocean circulation and its variability patterns, thus revealing the recurrent circulation patterns during different seasons.
McGuire et al.~\cite{mcguire2010spatiotemporal} developed a spatial neighborhood generation method combined with a temporal interval generation method to identify the regions with similar characteristics in SST data. 

Although there are already numerous studies on ST clustering, the physical meaning, and significance of generated clusters are sometimes debatable.
Moreover, there are no guarantees to find meaningful ST patterns because of the high spatial and temporal variability in ocean data. Therefore, how to develop an effective ST clustering method to produce clustering patterns with physical meanings is still a big problem for ocean science.


\begin{figure}[]
  \centering
\includegraphics[width=\textwidth]{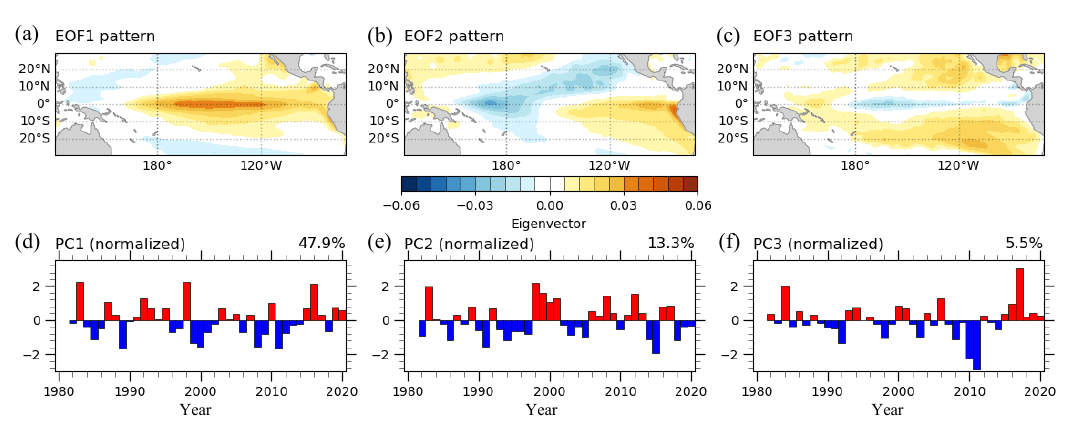}
  \caption{Spatial distributions (a,b,c) and time series (d,e,f) of the SST anomalies of the first three modes by EOF decomposition in the North Pacific Ocean from 1982 to 2020.}
    \label{article234355}
\end{figure}

\subsubsection{Empirical Orthogonal Function analysis} One of the most fundamental tools in ST pattern finding is Empirical Orthogonal Function (EOF) analysis. 
EOF is synonymous with the eigenvectors in the traditional eigenvalue decomposition of a covariance matrix~\cite{hannachi2007empirical}. 
As pointed out by Cressie and Wikle~\cite{cressie2015statistics}, in the discrete case, EOF analysis is simply principle component analysis (PCA). 
In the continuous case, it is a Karhunen-Loeve~(KL) expansion. 

EOF analysis has been traditionally used to identify a low-dimensional subspace that best explains the ST variance of ocean data. 
In order to better understand the spatial distribution characteristics of the SST in the North Pacific, the EOF method~\cite{article234355} is used to analyze the variability of SST in during 1980-2020, and three spatial distributions with large variance contribution were obtained, as shown in Fig.~\ref{article234355}.
To capture clear ST patterns, researchers seek to identify dominant spatial structures and their evolution over time. 
For instance, Harika and Virendra~\cite{rs13020265} utilized EOF to identify the three broad divisions of the ocean region, i.e., western, central, and eastern, based on the seasonal ice mass gain or loss from satellite data, which corresponds to the climatic changes.
Yan et al.~\cite{8684281} proposed an EOF-based method to retrieve the displacement signal for decorrelating targets in satellite data. 
 
In a word, EOF is a powerful analysis method that has been widely used in ocean science.
However, in the big data era, the existing versions of EOF are slightly out of date, and there is a need to develop new EOF-related methods 
to take full advantage of the large amount of ST ocean data and the advanced STDM techniques.  

\subsubsection{Spatial-temporal correlation mining}
Correlation mining in ocean aims to analyze the relationship between ocean variables, e.g., understanding ocean-atmosphere interaction, studying the effects of climate change, and finding the hidden correlations of teleconnection patterns. 
Various methods, including linear analysis methods and statistical decomposition methods, have been proposed for ST correlation mining in ocean.
Goldenberg and Shapiro~\cite{article22312r} used linear and partial linear correlations to analyze the relationship between the vertical wind shear in the Atlantic and the SST and Sahel rainfall patterns. 
Webster et al.~\cite{doi:10.1126/science.1116448} analyzed the linear correlations between basin-wide mean SST and seasonal Tropical Cyclone counts in all the major basins between 1970-2005 and concluded that the upward trend in Atlantic Tropical Cyclone seasonal counts cannot be attributed to the increased SST. 
Chen et al.~\cite{doi:10.1126/science.1209472} used the linear regression method to discover the correlations between different oceanic regions and fire activity in Amazon from the global SST data. 
Hoyos et al. \cite{doi:10.1126/science.1123560} used composite analysis to quantify how well one variable~(e.g., SST and humidity) explains the intensity in a typhoon event.
Yang et al.~\cite{yang2004trend} utilized singular value decomposition (SVD) to discover the correlations between the summertime cloud over Central-eastern China and the warming trend of SST in the Indian Ocean.
Liu et al.~\cite{liu2012tree} applied linear analysis to reveal the significant correlation between the forest resources and SST over the western-north Pacific Ocean area.
Liu et al.~\cite{https://doi.org/10.1002/2015JC011493} adopted the SOM method to reveal the patterns of the loop current system in the eastern Gulf of Mexico.

\subsection{Spatial-temporal Anomaly Detection}
Anomalies refer to the sets of observations that appear to deviate from expected behavior as compared with other observations.
ST anomaly detection in ocean domain is an important task and is used in a wide variety of ocean applications.
For example, trajectory anomaly detection can be used to detect whether a ship is currently undertaking a voyage that may cause an accident or whether it has changed its route to avoid bad weather conditions.
In this section, we summarize three typical ST anomaly detection tasks, i.e., trajectory anomaly detection, sensor time-series anomaly detection, and satellite image anomaly detection, for STDM in ocean.

\subsubsection{Trajectory anomaly detection}
%
During the voyage of the ship, various abnormal situations may be encountered, such as equipment failure, rocking and collision, and illegal activities. 
It is of high significance to implement effective trajectory anomaly detection, which enables timely detection and notification of abnormal events, and helps quickly take measures to protect property safety and people safety.
With over 1,490,776 ships tracking worldwide~\cite{wolsing2022anomaly},  it is difficult and time-consuming to manually find anomalies. 
Therefore, automatically analyzing the trajectory data to quickly and accurately detect anomalies is of high necessity. 
To this end, many trajectory anomaly detection methods have been proposed, and these methods can be roughly divided into two categories, i.e., rule-based methods, and learning-based methods. 

Rule-based methods use a set of rules for anomaly detection, which requires a good understanding of the anomalies~\cite{kazemi2013open, soleimani2015anomaly, riveiro2009interactive}. 
For example, Soleimani et al.~\cite{soleimani2015anomaly} proposed a near-optimal path method to detect ship trajectory anomalies and 
can also provide reasonable explanations for the detected anomalies. 
However, in practice, it is difficult to design a complete set of rules to cover various trajectory anomalies.
Moreover, rule-based methods are often time-consuming and labor-intensive, which also limits their applications.

\begin{figure}
	\centering
	\subfigure[All AIS tracks]{
		\begin{minipage}[b]{0.23\textwidth}
			\includegraphics[width=1\textwidth, height=\textwidth]{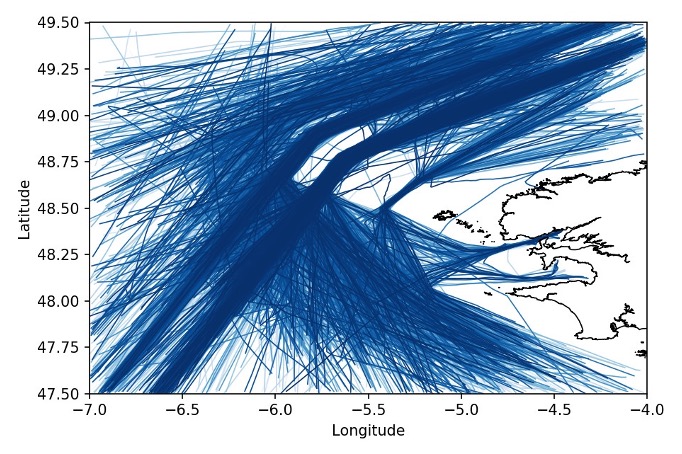} 
		\end{minipage}
	}
    	\subfigure[LSTM]{
    		\begin{minipage}[b]{0.23\textwidth}
   		 	\includegraphics[width=1\textwidth, height=\textwidth]{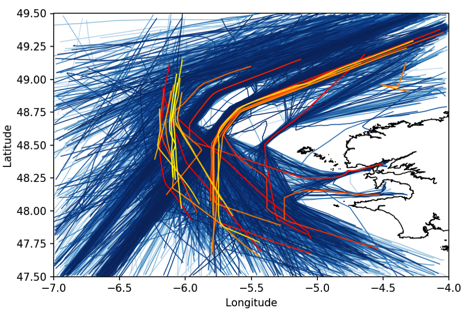}
    		\end{minipage}
    	}
	\subfigure[DBSCAN-based]{
		\begin{minipage}[b]{0.23\textwidth}
			\includegraphics[width=1\textwidth, height=\textwidth]{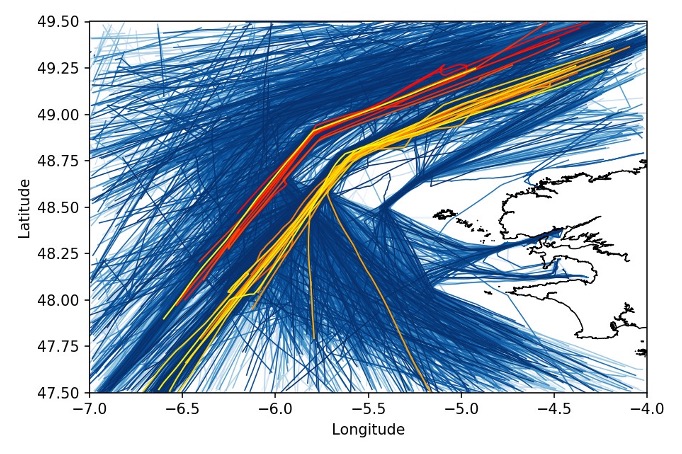} 
		\end{minipage}
	} 
    	\subfigure[GeoTrackNet]{
    		\begin{minipage}[b]{0.23\textwidth}
		 	\includegraphics[width=1\textwidth, height=\textwidth]{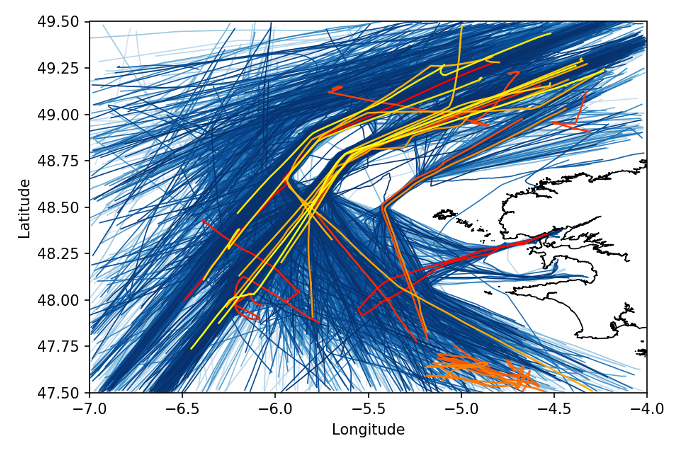}
    		\end{minipage}
    	}
	\caption{The visualization of AIS trajectories and the corresponding detection results of three trajectory anomaly detection methods~\cite{nguyen2021geotracknet} from January 1 to March 31, 2017, where blue color indicates normal trajectories and other colors indicate the detected abnormal trajectories.}
	\label{anaoly}
\end{figure}

For learning-based methods, many researchers utilize stochastic methods and machine learning methods, e.g., the classification model~\cite{fahn2019abnormal}, gaussian process~\cite{rong2019ship}, gaussian mixture model~\cite{anneken2015evaluation}, density estimation model~\cite{rong2020data}, and bayesian network~\cite{forti2021bayesian}, to achieve trajectory anomaly detection.
For example, Fahn et al.~\cite{fahn2019abnormal} proposed a typical classification method based on SVM and K-Nearest Neighbors~(KNN) to classify ship trajectories for detecting the abnormal ones.
However, classification-based detection methods require manually labeling the trajectories, and are prone to the risk of overfitting~\cite{han2020dbscan}.
To solve this problem, Han et al.~\cite{han2021modeling} introduced the density-based clustering method DBSCAN to learn the distribution of trajectories without labeling and achieved good performance.  
Xie et al.~\cite{10151936} developed the Gaussian Mixture Variational Autoencoder (GMVAE) using an unsupervised classification method and achieved a high detection rate~(91.26\%).
Martha et al.\cite{ferreira2022novel} combined autoregressive and cluster analysis to detect anomalies in maritime navigation trajectories.
To further capture the hidden and deep correlations, Duong et al. \cite{nguyen2021geotracknet} proposed GeoTrakNet which utilizes neural networks to learn the probability representation of AIS trajectories for detecting trajectory anomalies. 
Fig.~\ref{anaoly} visualizes the results of three typical trajectory anomaly detection methods on the same dataset, where the GeoTrakNet achieves the best performance.
Kexin et al.~\cite{li2023abnormal} proposed an adaptive vessel trajectory anomaly detection model based on transfer learning and transformer architecture, which introduces multiple attention modules to mine the dynamic dependencies between data points in the trajectory sequence.

Although there exist many methods for trajectory anomaly detection in ocean, an objective and fair comparison between these methods is hardly possible due to closed evaluation datasets and missing ground truth for anomalies.
Therefore, it is worthwhile to further explore unsupervised or semi-supervised trajectory anomaly detection methods based on the large volume of unlabeled trajectory data and limited labeled data. 

\subsubsection{Sensor time-series anomaly detection}
Many in-situ sensors are usually located in harsh environmental conditions and transmit the collected data across networks. Hence, the collected data usually contain a series of outliers and anomalies.
Detecting these outliers and anomalies is very important for improving the quality and reliability of the ST ocean datasets.
To this end, the major difficulty is to differentiate the true anomaly rather than the regular fluctuation, which requires a good understanding of the corresponding environmental circumstances.
Researchers have developed many automated statistical and machine-learning methods to jointly use spatial and temporal information for detecting sensor time-series anomalies.
For example, Barua~\cite{barua2007parallel} used a parallel wavelet transform to detect ST outliers in SST time-series data. 
Kumars~\cite{MahmoodiGhassemi+2018+44+50} utilized three typical outlier detection algorithms, i.e., Box-plot (BP), Local Distance-based Outlier Factor (LDOF), and Local Outlier Factor (LOF), to detect the outliers in wave height time-series records.
Anbaro~\cite{anbarouglu2009spatio} used an ST ARIMA model to define the coherent ST neighborhoods and then detected the outliers which have their values significantly different from the mean of nearby ST neighborhoods.
Wu et al.~\cite{DBLP:conf/kdd/WuLC08} detected the ST outliers in precipitation data by storing high discrepancy spatial regions over time in a tree. 
Ali et al.~\cite{9342827} utilized a robust z-score algorithm and isolation forest to detect outliers in sensor networks, achieving good detection accuracy with a low false alarm rate.
Cheng and Li~\cite{cheng2006multiscale} developed a multi-scale algorithm to detect abnormal coastal anomalies by evaluating the changes between ST scales.
Recently, Kim et al.~\cite{kim2021outlier} developed an autoencoder-based unsupervised learning method to detect outliers in SST data.

\subsubsection{Satellite image anomaly detection}
Satellite image anomalies refer to defective remote sensing images which may contain abnormal color, color cast, and unknown objects. Satellite image anomaly detection has a broad application, e.g., ship invading detection and extreme cloud detection.
Detecting image anomalies is more difficult than detecting the anomalies in trajectory data and time series data because visual features are more difficult to capture than numerical features.
Marino et al.~\cite{marino2016depolarization} proposed an incoherent dual-polarization detector to identify the anomaly of small icebergs of satellite images. Han et al.~\cite{han2021spectral} proposed a spectral anomaly detection method based on dictionary optimization with sparse and low-rank constraints. The error matrix that cannot be fully expressed by the optimized spectral dictionary is defined as anomalies.
 
Recently, various CNN-based methods have been applied for satellite image anomaly detection due to their high capability in analyzing image data and achieved promising detection results~\cite{rs11010047,rs13081506}.
For example, Wang et al.~\cite{rs11010047} proposed a spatial pyramid pooling net combining the multivariate Gaussian distribution to extract candidate regions of ships by regarding a ship as an abnormal ocean area. 
However, most satellite image anomaly detection methods still rely on the labeled images with manual expert annotations, and cannot take full advantage of the large amount of unlabeled satellite image data. 


%
\section{Future Opportunities}\label{sec:challenges}
The progress in STDM for ocean science has made significant impacts on various ocean applications as discussed above.
However, there still exist many open issues that require continuous efforts from both computer science and ocean science.  
In this section, we highlight some promising research directions for further advancing the techniques of ocean STDM.


\textbf{Fusing physical models with data-driven models.}
The physical models utilize a series of functions or theories to model the physical processes within the ocean~(e.g., ocean circulation and heat circulation) and are an important type of technique for various oceanic problems. 
In the past centuries, many classical physical ocean models, including ocean calculation models~\cite{haidvogel1999numerical, dijkstra2005nonlinear}, numerical ocean models~\cite{crowley1968global}, and physical oceanography models~\cite{stewart2008introduction}, have been developed. 
These methods only have a small number of factors determined based on expert knowledge but achieve good performance on many oceanic tasks.
Recently, data-driven models have received increasing attention and are of high ability to capture hidden information by utilizing historical data for model training~\cite{wangDeep2022}.
Some recent studies find that it is reasonable and feasible to combine physical models with data-driven methods~\cite{von2021informed}, which can produce more robust and accurate models that combine the advantages of both physics models and data-driven models.
Typical examples include physics-aware AI, physics-informed machine learning, and physics-guided machine learning~\cite{karniadakis2021physics,ahmad2020physics}.
Taking the SST prediction task as an example, Arka et al.~\cite{jia2021physics} integrate traditional physical laws (e.g., temperature density and energy conservation) and recurrent graph networks to predict SST. 
Compared with the traditional STDM methods, the physical-guided machine learning models can achieve high physical consistency and reduce the uncertainty of results.
However, there also exist multiple obstacles, e.g., the difference in data formats, problem definitions, and model structures, to fuse physical models with data-driven models. 
Therefore, in the future designing a general paradigm to fuse physical models with data-driven models is an interesting research direction.

\textbf{Fusing multi-source ocean datasets of different modalities.} As discussed in Section~\ref{sec:data}, the ocean data used in STDM tasks usually has multiple sources, including satellite data, in-situ data, ship data, and reanalysis data, and these datasets are in different modalities. 
Fusing multi-source datasets can obtain comprehensive information to make STDM models achieve better performance than using a single dataset or data of a single modality.
Although there are already some studies on data fusion (cf. Section~\ref{sec:datafusion}), there are still many challenges in fusing multi-source datasets. 
First, most existing ocean STDM methods focus on fusing different data sources for the same ocean factor (e.g., SST) in single tasks~(e.g., SST prediction), and cannot combine the observations of multiple factors~(e.g., SST, ocean current, and Chl-a) for multiple tasks~(e.g., SST prediction, current prediction, and Chl-a prediction) at the same time. 
Second, it is difficult to handle the heterogeneity in spatial and temporal resolutions for different datasets.
Furthermore, how to fuse the image satellite data, gridded reanalysis data, and ST points data is also a challenging problem. 
Recently, some researchers try to use transfer learning and meta-learning methods to fuse different ocean data sources and introduce the pre-training models in computer vision and urban computing~\cite{https://doi.org/10.1029/2021WR029579} to capture the features of multi-modal ocean datasets.
However, the challenges mentioned above are still not well addressed and require deeper investigation in the future.

\textbf{Improving the interpretability of deep STDM methods.} 
Many STDM approaches are based on deep-learning models, e.g., CNN, RNN, and Transformer, and achieve good performance on various tasks.
However, these approaches are typically regarded as black boxes with poor interpretability~\cite{fan2021interpretability}.
The interpretability of STDM methods means the ability to present and explain the model behaviors in understandable forms to humans, which enables users to easily make reasonable and convincing decisions~\cite{du2019techniques}. 
For ocean science, it is of high importance to build a connection between the STDM results and the underlying physics to provide valuable advice for various ocean applications, e.g., the warning of extreme weather conditions.
Although attention mechanisms have been used in some previous STDM studies for ocean science to increase the model interpretability~(e.g., periodicity and local spatial dependency), how to develop more interpretable STDM methods to reveal the natural laws of the ocean remains an open problem in this field~\cite{zhang2021survey}.

\textbf{Developing end-to-end STDM models.} Currently, many STDM methods for ocean science require that the input ST ocean data are complete and of high quality. To meet such requirements, data cleaning, data completion, and data fusion are conducted over the original ST ocean data. However, these processes are time-consuming and may lead to the loss of some crucial data information. For example, in the SST prediction task, the missing rate of SST records retrieved from the satellite data may be higher than 80\% for some periods and locations due to the cloud coverage~\cite{lian2023gcnet}.
Existing methods for SST prediction usually utilize data completion methods~(e.g., history average, optimal interpolation, and GAN-based networks) to fill in the missing values before conducting the prediction.
However, these methods ignore the sparse distribution of original data and the reasons behind the sparsity~(e.g., typhoons and heavy rainfall), which seriously damages the prediction accuracy.
Therefore, developing end-to-end STDM models to learn knowledge and patterns directly from the original sparse ST ocean data is a promising direction.   


\textbf{Designing large models for ocean science.} 
Recently, the large models~(e.g., BERT~\cite{devlin2018bert}, GPT~\cite{floridi2020gpt}, and ViT~\cite{ranftl2021vision}) have 
received tremendous popularity in computer science, and surpassed many state-of-the-art machine learning models by pre-training on big datasets and fine-tuning for different downstream tasks.
In ocean science, large ST ocean datasets of wide spatial coverage have been utilized to train STDM models for various oceanic tasks, which lays the solid foundation for designing unified large models.
For example, a recent big weather forecasting model~(i.e., Pangu-Weather~\cite{bi2023accurate}) is trained on 39-year global weather datasets and achieves great performance on various downstream tasks such as medium-range weather forecasting, extreme weather forecasting, and ensemble weather forecasting.
In fact, pre-training models trained on large ST ocean datasets can capture the underlying patterns and regularity more effectively and produce more accurate analyzing results for various oceanic tasks. 
Therefore, designing large models in the domain of ocean science is of high necessity to address the heterogeneity issues in data and tasks. 


\section{Conclusion}
The increasing numbers of satellites, buoys, and other ocean observation facilities have generated a flood of ST ocean data, which makes it possible to develop STDM methods for various ocean applications, e.g., weather forecasting and typhoon detection.
The past decades have witnessed the rapid development of STDM techniques for ocean science, which enable us to uncover the underlying ST dependencies within ocean data for solving the challenging problems in ocean science.
This survey provides a comprehensive overview of the existing ocean STDM studies, including the publically available ST ocean datasets and their unique characteristics as well as the data visualization tools, the data quality enhancement methods, and the methodologies for different types of STDM tasks. In addition, some promising research opportunities for future research are also pinpointed.    
We hope that this paper can help computer scientists identify important research issues of ocean big data, and inspire researchers in ocean science to apply advanced STDM techniques to a variety of ocean applications.
\bibliographystyle{ACM-Reference-Format}
\bibliography{base}


\end{document}